\pdfoutput=1

\documentclass[11pt]{article}

\usepackage[final]{acl}

\usepackage{times}
\usepackage{latexsym}
\usepackage{booktabs}

\usepackage[T1]{fontenc}

\usepackage[utf8]{inputenc}

\usepackage{microtype}

\usepackage{inconsolata}

\usepackage{graphicx}

\usepackage{subcaption}

\usepackage{MnSymbol}

\makeatletter
\DeclareRobustCommand*{\escapeus}[1]{%
  \begingroup\@activeus\scantokens{#1\endinput}\endgroup}
\begingroup\lccode`\~=`\_\relax
   \lowercase{\endgroup\def\@activeus{\catcode`\_=\active \let~\_}}
\makeatother

\usepackage{tabularray}
\usepackage{cellspace}
\newcommand\cincludegraphics[2][]{\raisebox{-0.2\height}{\includegraphics[#1]{#2}}}
\usepackage{setspace}

\newcommand{\myemph}[1]{\textsf{{\escapeus{#1}}}}
\newcommand{\githublink}[2]{
    \hspace{-15pt}
    \begin{tblr}{
      colspec = {Q[c,m]Q[c,m]},
      stretch = 0,
      columns = {colsep=1.5pt},
      rows = {rowsep = 0pt},
    }%
     \cincludegraphics[width=1.1em, keepaspectratio]{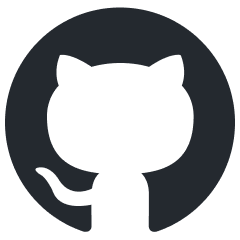} 
     & \setstretch{.5}\href{#1}{\small\myemph{#2}}\\
    \end{tblr}%
}

\newcommand{\huggingfacelink}[2]{
    \hspace{-15pt}
    \begin{tblr}{
      colspec = {Q[c,m]Q[c,m]},
      stretch = 0,
      columns = {colsep=1.5pt},
      rows = {rowsep = 0pt},
    }%
      \cincludegraphics[width=1.1em, keepaspectratio]{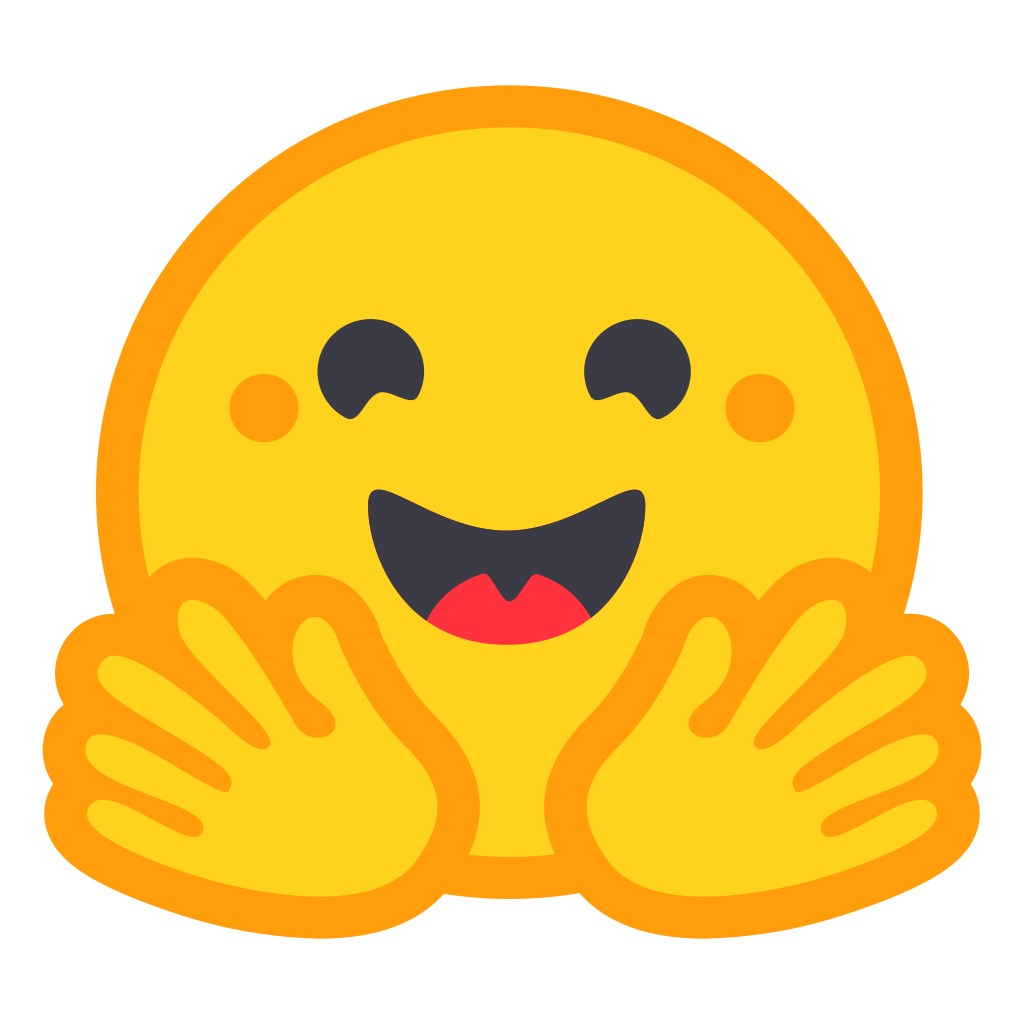}
      & \setstretch{.5}\href{#1}{\small\myemph{#2}}\\
    \end{tblr}%
}

\usepackage{mymacros}

\author{
Jacob Mitchell Springer{$^\dagger$} \quad \textbf{Vaibhav Adlakha}{$^\ddagger$} \quad \textbf{Siva Reddy}{$^{\ddagger, \S}$} \\ \textbf{Aditi Raghunathan}{$^\dagger$} \quad \textbf{Marius Mosbach}{$^\ddagger$} \\
$^{\dagger}$~\text{Carnegie Mellon University} \\
$^{\ddagger}$~\text{Mila -- Quebec AI Institute, McGill University} \\
$^{\S}$~\text{Canada CIFAR AI Chair} \\
\footnotesize{\texttt{jspringer@cmu.edu} \quad \texttt{marius.mosbach@mila.quebec}}}

\usepackage{todonotes}
\usepackage[normalem]{ulem}

\makeatletter
\newcommand*\iftodonotes{\if@todonotes@disabled\expandafter\@secondoftwo\else\expandafter\@firstoftwo\fi}  %
\makeatother

\renewcommand{\R}{\mathcal{R}}

\usepackage{soul}
\usepackage{color}

\title{Understanding the Influence of Synthetic Data for Text Embedders} 

\begin{document}

\maketitle

\begin{abstract}
Recent progress in developing general purpose text embedders has been driven by training on ever-growing corpora of synthetic LLM-generated data. Nonetheless, no publicly available synthetic dataset exists, posing a barrier to studying its role for generalization. To address this issue, we first reproduce and publicly release the synthetic data proposed by \citet{wang-etal-2024-improving-text} (Mistral-E5). Our synthetic data is high quality and leads to consistent improvements in performance. Next, we critically examine where exactly synthetic data improves model generalization. Our analysis reveals that benefits from synthetic data are sparse and highly localized to individual datasets. Moreover, we observe \emph{trade-offs} between the performance on different categories and data that benefits one task, degrades performance on another. 
Our findings highlight the limitations of current synthetic data approaches for building general-purpose embedders and challenge the notion that training on synthetic data leads to more robust embedding models across tasks.
\vspace{0.5em}\\
\githublink{https://github.com/jakespringer/open-synthetic-embeddings}{jakespringer/open-synthetic-embeddings}\vspace{0.2em}\\
\huggingfacelink{https://huggingface.co/datasets/jspringer/open-synthetic-embeddings}{jspringer/open-synthetic-embeddings}
\end{abstract}

\section{Introduction}
\label{sec:introduction}

Mirroring the success of generative LLMs \citep{ouyang2022training, jiang2023mistral7b, grattafiori2024llama3herdmodels}, the NLP community has invested in building general-purpose embedding models---single models capable of producing embeddings for a wide array of embedding tasks, spanning classification, clustering, retrieval, reranking, and text-similarity estimation \citep{li-etal-2023-gte, wang-etal-2024-improving-text, springer2024repetition, behnamghader-2024-llm2vec, muennighoff-etal-2024-generative, lee-etal-2024-nvembed}.

\begin{figure}[t]
    \centering
    \includegraphics[width=0.99\columnwidth]{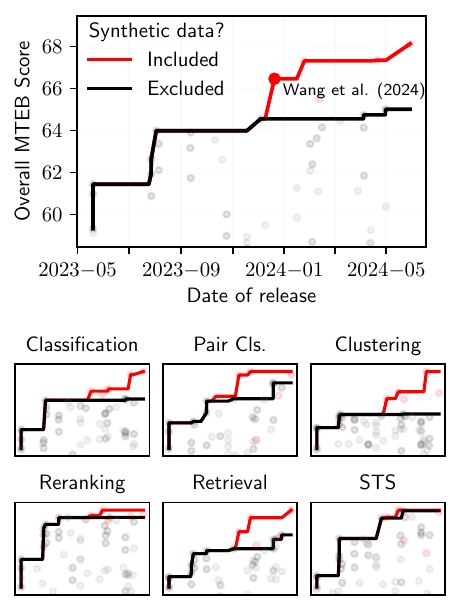}
    \caption{Performance on MTEB across time. Starting with \citet{wang-etal-2024-improving-text}, models trained on synthetic data have led to considerable improvements on the MTEB leaderboard. We exclude more recent models that were trained on in-domain data. }
    \label{fig:teaser}
\end{figure}

Beyond architectural innovations and progress in base models, much of the recent progress on general-purpose embedders can be attributed to training on synthetic training data—for example, by leveraging GPT-4 to produce synthetic data that expands existing training datasets for embeddings to new tasks (see Figure \ref{fig:teaser}) \citep{wang-etal-2024-improving-text}. Leveraging synthetic data is based on the premise that LLMs generate more diverse and high-quality data compared to human-annotated datasets, which are often limited in size.

In our work, we critically examine the implicit assumption of this paradigm that training on synthetic data will broadly improve the general-purpose quality of the model. We approach this by training models on different compositions of synthetic data to estimate the influence of different commonly adopted types of synthetic data on the downstream performance of the model. Strikingly, \textbf{we find that training on synthetic examples designed for a particular task can degrade the performance of other tasks, challenging the notion that training on more diverse synthetic data is strictly better.} Moreover, we observe that synthetic data leads to sparse improvement across tasks, showing no statistically significant improvement on a majority of MTEB tasks.

To conduct our analysis, we reproduce the synthetic data from \citet{wang-etal-2024-improving-text} and release this data publicly. In our reproduction, we compare the effectiveness of LLaMA-3.1-8B and LLaMA-3.1-70B for generating synthetic data, both of which are more cost-efficient than the GPT-family models used by \citet{wang-etal-2024-improving-text} and are publicly available \citep{grattafiori2024llama3herdmodels}. Our results show that synthetic data generated by LLaMA-3.1-8B performs nearly as well as data from the larger 70B model, while costing $5\times$ less than the 70B model and over $50\times$ less than GPT-4o.

In total, our results underscore the need to develop broad and robust methods for generating and training on synthetic data that do not exhibit trade-offs, and we hope that our public release of synthetic training data will accelerate open research into the development and understanding of general-purpose embedding models.

\section{Generating high quality synthetic data}
\label{sec:generating_high_quality_synthetic_data}

\citet{wang-etal-2024-improving-text} have demonstrated that the addition of large quantities of LLM generated synthetic data can led to substantial improvements in embedding quality. In fact, synthetic data has been so successful that the current gold standard text embedding benchmark---MTEB \citep{muennighoff-etal-2023-mteb}---is largely dominated by models that train, at least in part, on synthetic data \citep{lee-etal-2024-nvembed, muennighoff-etal-2024-generative, SFRAIResearch2024}. 

However, the synthetic datasets used to train these models are typically generated using proprietary LLMs and remain unavailable to the scientific community, making it difficult to understand their role for the generalization of general-purpose embedding models. 

We address this by reproducing and publicly releasing the synthetic data from our replication of \citet{wang-etal-2024-improving-text}. We describe our approach for generating synthetic data as follows.

\paragraph{Dataset generation pipeline.}

We generate data following the pipeline proposed by \citeauthor{wang-etal-2024-improving-text} which offers an effective method to generate synthetic data that has diverse structure and content.
More specifically, we generate data of six different categories based on query and document length: \textit{short-short, long-long, short-long, long-short, bitext, and STS}. Short examples, such as queries in short-long---consist of a few words or a single sentence, while long examples comprise multiple sentences. In short-short, long-long, STS, and bitext pairs, queries and documents are drawn from the same distribution, modeling semantic similarity estimation. In contrast, long-short and short-long pairs involve different distributions, covering tasks such as classification and retrieval. 
To ensure diversity of content, we generate data in two steps: 

\begin{enumerate}
    \item \textit{Brainstorming:} We generate task descriptions for each of the categories outlined above.
    \item \textit{Instance generation:} We generate training examples based on both the output of the brainstorming stage and the associated category. Each training example consists of an instruction, a query, a positive example relevant to the query, and a hard negative example\footnote{In some cases, the hard negative is mined instead of generated. We refer to \Cref{appendix:sec:synthetic_data_generation} for details.} that is only superficially relevant to the query.
\end{enumerate}

\noindent Following the pipeline described above, we generate approximately 500k synthetic examples from LLaMA-3.1-8B and LLaMA-3.1-70B, respectively. More details of the generation process along with prompts, examples, and the final composition of synthetic data are in  \Cref{appendix:sec:synthetic_data_generation}.

\paragraph{Training and evaluation setup.}

We experiment with three different models: Mistral-v0.1-7B, Mistral-v0.2-7B \citep{jiang2023mistral7b}, and Qwen2-1.5B \citep{yang2024qwen2technicalreport}, and compare training without synthetic data to training with synthetic data sourced from either Llama-3.1-70B or Llama-3.1-8B. We follow \citet{wang-etal-2024-improving-text} and mix our synthetic data with the public E5 dataset, using the version released by \citet{springer2024repetition}.\footnote{\citet{springer2024repetition} is the only work that replicates the dataset curation of \citet{wang-etal-2024-improving-text} and publicly releases it.} 

We evaluate on the Massive Text Embedding Benchmark (MTEB; \citealt{muennighoff-etal-2023-mteb}), which consists of 56 embedding datasets spanning seven different tasks. For a full list of tasks and datasets along with the details of our evaluation, see \Cref{appendix:sec:mteb_details}.

\begin{figure}[t]
    \centering
    \begin{subfigure}[t]{0.48\textwidth}
        \centering
        \includegraphics[width=\textwidth,trim={0.24cm 0.24cm 0.24cm 0.24cm},clip]{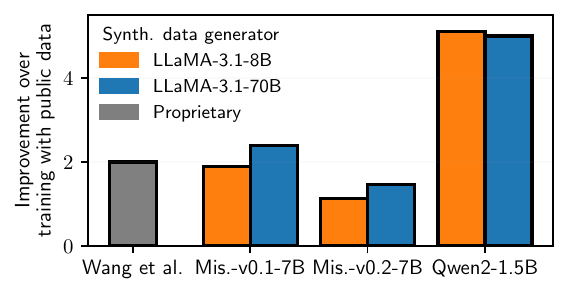}
        \label{fig:placeholder-a}
    \end{subfigure}%
    \vspace{-1em}
    \caption{Improvement in MTEB score when adding synthetic data to the training mixture. Across all settings, our results are consistent with \citet{wang-etal-2024-improving-text}, showing that training on synthetic data leads to higher MTEB performance.}
    \label{fig:comparison-with-mistral-e5}
\end{figure}

\paragraph{Synthetic data leads to improvements on MTEB.} 
We compare our synthetic data to \citet{wang-etal-2024-improving-text}, and \citet{chen2024little}---a method involving generating synthetic data from smaller language models that are fine-tuned to produce high quality synthetic data.

As shown in \Cref{table:synthetic-data-results} and \Cref{fig:comparison-with-mistral-e5}, our synthetic data replicates the overall relative improvement of \citet{wang-etal-2024-improving-text} as well as \citet{chen2024little}. We note that we do not match the exact scores of \citet{wang-etal-2024-improving-text}, likely due to differences in the hyperparameters and general setup used for training. However, in all cases our synthetic data substantially improves performance. In fact, our Mistral-v0.2-7B model trained on LLaMA-3.1-70B synthetic data outperforms the scores reported by \citet{wang-etal-2024-improving-text}, and \citet{chen2024little} reaching an absolute score of 66.9. It is noteworthy that synthetic data especially benefits Qwen2-1.5B ($+5.1$ points compared to $+2.4$ points for Mistral-v0.1-7B).

\section{Investigating task influence}
\label{sec:experiments}

Having replicated the findings of \citet{wang-etal-2024-improving-text}, i.e, that synthetic data improves MTEB performance on average, we now turn to studying to what extent the different task types of the synthetic data influences performance of each category of MTEB. More specifically, does each type of synthetic data broadly improve performance, or are improvements localized to particular tasks? Even more importantly, are there inherent performance \emph{trade-offs} between different tasks based upon the exact synthetic data composition? Having answers to these questions is crucial for our understanding on how synthetic data impacts the overall generalization of text embedding models.

\paragraph{Estimating data influence.}

\begin{table}[t]
    \centering
    \label{tab:model_scores}
    \begin{tabular}{lrrr}
    \toprule
    \textbf{Base model}       & \textbf{Pub. data}   & \textbf{L-8B} & \textbf{L-70B} \\
    \midrule
    Mistral-v0.1-7B  & 63.1   & 65.0         & 65.5         \\
    Mistral-v0.2-7B  & 65.4  & 66.6        & \textbf{66.9}         \\
    Qwen2-1.5B    & 58.3  & 63.4         & 63.3         \\ \midrule
    \citeauthor{wang-etal-2024-improving-text}   & 64.6   & \multicolumn{2}{c}{66.6$^*$} \\
    \citeauthor{chen2024little} & & \multicolumn{2}{c}{66.5$^*$} \\
    \bottomrule
    \end{tabular}
    \caption{MTEB scores for different base models and synthetic data generators. In all cases, adding synthetic data to the training mixture leads to improved performance. Note that \citeauthor{wang-etal-2024-improving-text} and \citeauthor{chen2024little} use the GPT model family for at least part of their synthetic data generation. See Appendix~\ref{app_tab:full_mteb} for more detailed evaluations.}
    \label{table:synthetic-data-results}
\end{table}

To address the questions posed above, we measure how each synthetic subset affects model performance with an \textit{influence function}, which estimates the typical improvement that a specific subset of the training data contributes to the final performance. We consider specifically four of the synthetic data categories: short-short, short-long, long-long, and long-short. We train models with all possible $2^4=16$ combinations of this synthetic data, with the addition of a non-synthetic base dataset $\mathcal{D}$. To estimate the influence of a specific category $\mathcal{S}$, we split the 16 models into two groups: one group $\mathcal{P}^+_i$ of the eight models with training data that includes $\mathcal{S}$, and another group $\mathcal{P}^-_i$ of the eight remaining models that exclude $\mathcal{S}$ from the training data. We measure the influence of $\mathcal{S}$ by computing the difference between the mean performance of each group: $\mathbb{E}_{P \in \mathcal{P}^+_i}[\text{perf}(\mathcal{D} \cup P)] - \mathbb{E}_{P \in \mathcal{P}^-_i}[\text{perf}(\mathcal{D} \cup P)]$. This difference quantifies the improvement we expect to observe by training on each synthetic data category.

In addition to computing influence functions, we use a two-sided $t$-test to determine whether each synthetic category has a statistically significant (non-zero) contribution to MTEB performance.

\paragraph{Training and evaluation setup.}

We run experiments with Mistral-v0.2-7B and Qwen2-1.5B, and compare training on synthetic data from Llama-3.1-8B, and Llama-3.1-70B. For training, we adopt the same setup as \Cref{sec:generating_high_quality_synthetic_data}. Evaluation also follows \Cref{sec:generating_high_quality_synthetic_data} (MTEB) with the caveat that retrieval datasets are replaced with their faster versions. We refer the reader to \Cref{appendix:sec:training_details} for full details.

\paragraph{Cross-category generalization is sometimes negative.}

\begin{figure}[t]
    \centering
    \begin{subfigure}[t]{0.48\textwidth}
        \centering
        \includegraphics[width=\textwidth]{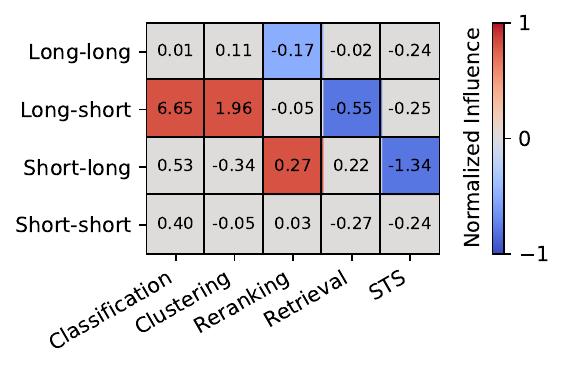}
        \label{fig:cross-category-results}
    \end{subfigure}
    \vspace{-1em}
    \caption{Influence of different training data categories on the MTEB evaluation categories when training Mistral-v0.2-7B on synthetic data from Llama-3.1-70B. Colored cells indicate statistically significant influence ($p<0.05$). The color indicates the normalized influence: influence is rescaled so that the maximum (absolute) influence has a value of $\pm 1$. Additional results in Appendix~\ref{app_tab:full_mteb}.}
    \label{fig:category-ablations}
\end{figure}

\Cref{fig:category-ablations} plots the influence of each synthetic data set on each task category for Mistral-v0.2-7B using synthetic data from Llama-3.1-70B. Surprisingly, we find a trade-off between the performance of different categories. For example, training on the synthetic long-short dataset benefits classification and clustering performance by $6.65$ and $1.19$ points on average. Similarly, short-long improves reranking performance ($+0.27$ points) but harms sentence similarity ($-1.34$ points). We observe similar trade-offs when using Qwen2-1.5B as a base model, and when training with LLaMA-3.1-8B synthetic data.

\paragraph{Synthetic data improves performance sparsely.}

We indicate the statistical significance of each influence estimate in \Cref{fig:category-ablations}. Often, the synthetic data has \emph{no} statistically significant influence on many of the MTEB evaluation categories. In fact, short-short data has no statistically significant improvement for any MTEB category.

\begin{figure}[t]
    \centering
    \begin{subfigure}[t]{0.48\textwidth}
        \centering
        \includegraphics[width=\textwidth]{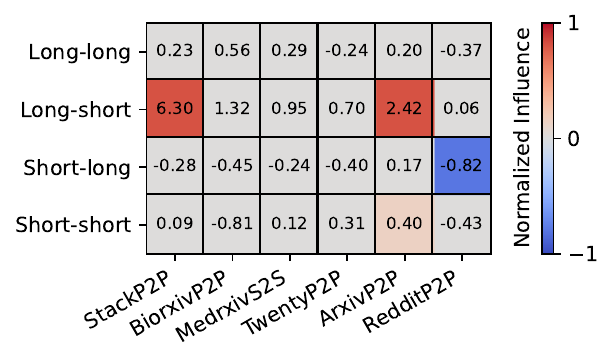}
        \label{fig:cross-dataset-results}
    \end{subfigure}
    \vspace{-1em}
    \caption{Influence of different training data categories on MTEB clustering tasks when training Mistral-v0.2-7B on synthetic data from Llama-3.1-70B. Colored cells indicate statistically significant influence ($p<0.05$). The color indicates the normalized influence: influence is rescaled so that the maximum (absolute) influence has a value of $\pm 1$. Additional results in Appendix~\ref{app_tab:full_mteb}.}
    \label{fig:dataset-ablations}
\end{figure}

Even within a particular evaluation category that improves with synthetic data, the majority of tasks observe no statistically significant improvement with synthetic data. For example, we observe that the long-short dataset, which generally improves the performance of clustering tasks, only improves StackExchangeClusteringP2P and ArxivClusteringP2P with significance (see \Cref{fig:dataset-ablations}). 

\section{Related work}
\label{sec:rlated_work}

\paragraph{General-purpose text encoders.}

A major challenge when training embedding models is that the notion of similarity these models learn is corpus-driven and hence, models often fail to generalize beyond the similarity definitions they saw during contrastive training \citep{thakur2021beir, muennighoff-etal-2023-mteb, ravfogel-etal-2024-descriptionbased}.

Driven by benchmarks such as BEIR \citep{thakur2021beir} and MTEB \citep{muennighoff-etal-2023-mteb}, the community has shifted its focus on building text embedding methods that generalize to multiple tasks and domains. While prior BERT-based approaches relied on complex multi-stage pipelines to achieve this goal \citep[\textit{inter alia}]{li-etal-2023-gte, shitao-etal-2023-bge}, recent approaches which are based on decoder-only LLMs have shown superior performance \citep[\textit{inter alia}]{wang-etal-2024-improving-text, behnamghader-2024-llm2vec, springer2024repetition, muennighoff-etal-2024-generative, lee-etal-2024-nvembed}. Leveraging advances in instruction-following capabilities of LLMs, these models achieve improved generalization to novel tasks and domains by using natural language instruction combined with multi-task contrastive learning.

\paragraph{Synthetic data for text embeddings.}

Synthetic data has been previously used in the information retrieval literature to generate pseudo queries or hypothetical documents \citep[\textit{inter alia}]{nogueira2019document, dai2023promptagator, wang-etal-2023-query2doc}. Recently, \citet{wang-etal-2024-improving-text} extended this paradigm to general-purpose text embedding methods, by using LLMs to generate high-quality training data for diverse embedding tasks. Since then, synthetic data generation has become a widely adopted strategy for improving text embedding models, particularly for models competing on the MTEB leaderboard.

\section{Conclusion}
\label{sec:conclusion}

In this work, we investigate how training on %
synthetic data 
influences downstream text embedding tasks. Our experiments confirm previous findings that synthetic data can significantly boost overall performance on MTEB, but offer a more nuanced perspective: we find that training on certain synthetic data categories can exhibit trade-offs in the task-specific performances. For example, while some synthetic data categories improve classification or clustering, they may degrade retrieval performance. Our results highlight that the addition of synthetic data does not always strictly improve text embedding models. Instead, its benefits are nuanced, requiring refined generation and training strategies to balance trade-offs. 

Beyond our analysis, we contribute a high quality reproduction of the synthetic data of \citet{wang-etal-2024-improving-text} which we release publicly, along with code for our reproduction. By releasing our data and code, we aim to support further research into optimizing synthetic training for general-purpose text embeddings.

\section*{Limitations}
\label{sec:limitations}

\paragraph{Role of the base dataset.}

In our experiments, we follow previous work \citep{wang-etal-2024-improving-text, muennighoff-etal-2024-generative} and combine synthetic data with an existing mixture of publicly available datasets \citep{springer2024repetition}. There might be some non-trivial interactions between our synthetic dataset and the existing data which we did not control for in our setup. We hope that releasing our synthetic data will encourage future work to explore potential interactions between the base dataset and the synthetic data added to it. 

\paragraph{Synthetic data generated from more capable models.}

We cannot rule out the possibility that synthetic data generated from more capable LLMs might lead to different conclusions from ours. However, as with the role of the base dataset discussed above, we hope that by releasing our data to the community, we make it easy for future work to compare to our data and investigate potential properties of synthetic data that are responsible for (larger) improvements in generalization.

\section*{Acknowledgments}

Marius Mosbach is supported by the Mila P2v5 grant and the Mila-Samsung grant. Siva Reddy is supported by the Canada CIFAR AI Chairs program and the NSERC Discovery Grant program. 

This material is based upon work supported by the National
Science Foundation Graduate Research Fellowship under
Grant No. DGE2140739. Any opinion, findings, and conclusions or recommendations expressed in this material are
those of the authors(s) and do not necessarily reflect the
views of the National Science Foundation.

We gratefully acknowledge support from ServiceNow, Apple, NSF
and the AI2050 program at Schmidt Sciences (Grant
\#G2264481).

\bibliography{custom}

\appendix

\clearpage
\section{Background}
\label{sec:appendix:background}

Let $\vocab$ be a finite vocabulary of tokens. A \defn{sequence} $\sequence \in \seqspace$ is a finite concatenation of tokens $\mathbf{\sequence} = (\sequence_1, \ldots, \sequence_{|\sequence|})$ where each $\sequence_i \in \vocab$. Our goal is to train an \textbf{embedder} $\embed: \seqspace \to \R^d$, parameterized by $\theta \in \R^p$, which maps sequences to embeddings. 
The ideal embedder accurately estimates the similarity between any pair of examples $\mathbf{\sequence}_1, \mathbf{\sequence}_2 \in \seqspace$, parameterized by a metric of similarity between the embeddings. We follow common practice in the embedding literature and use cosine similarity as to estimate the similarity between examples $\hat{f}(\mathbf{\sequence}_1, \mathbf{\sequence}_2) = \cos(\embed(\mathbf{\sequence}_1), \embed(\mathbf{\sequence}_2))$ \citep[\textit{inter alia}]{li-etal-2023-gte, wang-etal-2024-improving-text, springer2024repetition, behnamghader-2024-llm2vec, muennighoff-etal-2024-generative}.

\paragraph{Contrastive learning.}

We train a general-purpose embedding model via contrastive learning on a dataset $\mathcal{D} = \{ \tuple{\sequence_i, \seqpos_i, \seqneg_i} \}_{i=1}^n$. Below we state the loss for a single instance $\tuple{\sequence_i, \seqpos_i, \seqneg_i}$:

\vspace{-1.25em}
\begin{align}
    & \mathcal{L}(\tuple{\sequence_i, \seqpos_i, \seqneg_i}) \\ 
    &= -\log \frac{\exp(\hat{f}(\sequence_i, \seqpos_i))}{\exp(\hat{f}(\sequence_i, \seqpos_i)) + \exp(\hat{f}(\sequence_i, \seqneg_i))}\quad. \nonumber
\end{align}

For simplicity, we stick to a single negative example above but note that in practice, we typically consider multiple negative examples per instance. In that case, the second term in the denominator becomes $\sum_j \exp(s(\sequence_i, \seqneg_{i,j}))$, where we take a sum over all negative examples for a given input $\sequence_i$.

\section{Synthetic data generation}
\label{appendix:sec:synthetic_data_generation}

\begin{table*}[ht!]
    \centering
    \begin{tabular}{lrrrrrr|r}
    \toprule
    \textbf{Model}       & \textbf{short-short}   & \textbf{short-long} & \textbf{long-long} & \textbf{long-short} & \textbf{bitext} & \textbf{STS} & \textbf{Total}\\
    \midrule
    LLaMA-3.1-8B  & 19,769   & 146,717   & 17,344  & 106,577   & 88,228   & 99,612 & 478,247\\
    LLaMA-3.1-70B  & 19,932   & 153,934   & 19,236  & 108,487   & 89,611   & 99,791 & 490,991 \\
    \bottomrule
    \end{tabular}
    \caption{Composition of synthetic data from LLaMA-3.1-8B and LLaMA-3.1-70B across different categories.}
    \label{tab:synthetic_data_composition}
\end{table*}

We follow the synthetic data generation pipeline of \citet{wang-etal-2024-improving-text}. We follow their prompt template for both brainstorming and instance generation. The composition of synthetic data for both LLaMA-3.1-8B and LLaMA-3.1-70B across different categories in detailed in Table~\ref{tab:synthetic_data_composition}. We also provide an example of short-short category sample in \Cref{tab:app_synthetic_data_example_short_short}, short-long in \Cref{tab:app_synthetic_data_example_short_long}, long-short in \Cref{tab:app_synthetic_data_example_long_short}, and long-long in \Cref{tab:app_synthetic_data_example_long_long}.

\section{Training details}
\label{appendix:sec:training_details}

\citet{wang-etal-2022-e5} were among the first to demonstrate that a powerful decoder‑only LLM can be transformed into a high‑quality text encoder. To obtain text embedding, they appended an [EOS] token to each input and constructed an embedding form its last layer representation. They fine-tuned on sentence-pair data using a contrastive learning objective (see \Cref{sec:appendix:background}) to ensure that the resulting text representations effectively capture the semantic content of the input text.
\citeauthor{wang-etal-2022-e5} trained Mistral-7B (\texttt{Mistral-7B-v0.1}, \citealt{jiang2023mistral7b}) on 1.8 million sentence pairs. 
They perform parameter-efficient fine-tuning via LoRA \citep{hu2022lora}, using a batch size of 2048. The entire fine‑tuning process takes roughly 18 hours on 32 V100 GPUs.

Our training procedure largely follows \citeauthor{wang-etal-2024-improving-text}, but we make minor modifications inspired by subsequent work. First, we use a more recent base model from the same model family (\texttt{Mistral-7B-Instruct-v0.2}). Next, following \citet{behnamghader-2024-llm2vec} and \citet{muennighoff-etal-2024-generative}, we enable bidirectional connections within the model architecture and employ mean pooling over the token embeddings instead of relying on the final [EOS] token representation. 

For the public portion of \citeauthor{wang-etal-2024-improving-text}'s training data, we use the replication provided by \citet{springer2024repetition}. This data consists of about 1.5M samples. For the synthetic portion, we generate about $\sim$500,000 samples following the methodology described in \Cref{sec:generating_high_quality_synthetic_data}.

We train the models with LoRA $r=16$ and $\alpha=16$ using a batch size of 2048. We use a maximum sequence length of 512 tokens for fair comparison to previous approaches. We use the AdamW optimizer with a learning rate of $4e-4$, linear learning rate warm-up for the first 100 steps, and weight decay with 0.1 coefficient afterwards. We train all models for one epoch. Training Mistral-7B on public + synthetic data ($\sim$2M samples) takes about 16 hours on 8 H100 GPUs. We will release the dataset, pre-trained models and the training code upon publication.

\section{MTEB evaluation details}
\label{appendix:sec:mteb_details}

Text embeddings have been widely used in various NLP tasks, however, traditionally the evaluation of text embeddings has been limited to a small set of datasets from a single task such as semantic textual similarity or text retrieval \cite{karpukhin-etal-2020-dense, wang-etal-2021-tsdae-using}, making it difficult to estimate generalization of the proposed methods.

To address this issue, \citet{muennighoff-etal-2023-mteb} proposed MTEB -- Massive Text Embedding Benchmark -- a single comprehensive evaluation suite that spans a total of 56 datasets across 7 distinct tasks (retrieval (15), reranking (4), classification (12), clustering (11), pair classification (3), semantic textual similarity (STS, 10), and summarization (1). The individual dataset sizes vary widely: STS datasets range from 1K to 20K pairs, classification datasets range from 500-5000 samples, and retrieval datasets such as MS MARCO \citep{bajaj-etal-2018-MSMARCO} include thousands of queries with a 6M document corpus.

In MTEB, every task is reformulated as an embedding task where the only requirement is that the model produces a vector (embedding) for each text input. For example, in classification, MTEB uses the embeddings as fixed features and trains a lightweight linear classifier (typically logistic regression) on top. The performance of the classifier is treated as a proxy for the quality of the embeddings. In clustering, embeddings are fed into standard clustering algorithms (like mini-batch k-means) to group similar texts. Retrieval, reranking and STS follow standard evaluation protocol is which embeddings of pairs of text are compared using cosine similarity. 

The unified casting of diverse tasks into an embedding framework, simplicity of use, open-source evaluation code\footnote{\href{https://github.com/embeddings-benchmark/mteb}{https://github.com/embeddings-benchmark/mteb}}, and a public leaderboard\footnote{\href{https://huggingface.co/spaces/mteb/leaderboard}{https://huggingface.co/spaces/mteb/leaderboard}} has led to widespread use of this benchmark within the NLP community, making it the de facto standard for evaluating text embedding models. 

\paragraph{Faster version of MTEB.}

One drawback of MTEB is that evaluating a model is very computationally intensive. 
This is largely due to the retrieval task category, where each dataset has separate corpus containing millions of documents. To address this, \citet{enevoldsen2025mmteb} developed smaller versions of the retrieval datasets contained in MTEB by carefully selecting candidate documents for each query in the dataset. They showed that keeping only 250 documents per query, selected via hard-negative mining, maintains the  absolute scores and model ranking compared to evaluating on the original datasets. While we use original datasets for evaluation in \Cref{sec:generating_high_quality_synthetic_data} for a fair comparison with \citet{wang-etal-2022-e5}, our analysis in \Cref{sec:experiments} uses the faster version of the retrieval datasets.

\section{Additional evaluations}
\label{app:additional_evals}

\subsection{Full MTEB evaluations}

We evaluate each model that we consider in the main paper on MTEB in Table~\ref{table:synthetic-data-results}. In this section, we expand these results by plotting the average scores for each of the categories in Table~\ref{app_tab:full_mteb}. We find that our synthetic data improves clustering performance most substantially.

\begin{table*}[t]
\begin{tabular}{llcccccccc}
\toprule
{} &  &  \textbf{Cls.} &  \textbf{Clust.} &  \textbf{P. Clas.} &  \textbf{Rera.} &  \textbf{Retr}. &  \textbf{STS} &  \textbf{Sum.} &  \textbf{Mean} \\
\multicolumn{2}{l}{\# of datasets $\rightarrow$} & 12 & 11 & 3 & 4 & 15 & 10 & 1 & 56 \\
\midrule
Qwen2-1.5B & Public data only &               70.2 &            42.3 &                    83.1 &           51.6 &           48.2 &     76.2 &               31.4 &        58.3 \\
Qwen2-1.5B & LLaMA-3.1-8B &                76.6 &            47.0 &                    86.2 &           55.5 &           54.1 &     80.2 &               30.8 &        63.4 \\
Qwen2-1.5B & LLaMA-3.1-70B &                76.5 &            47.2 &                    86.9 &           55.0 &           54.4 &     79.3 &               29.7 &        63.3 \\
\midrule
Mis.-v0.1-7B & Public data only &                75.6 &            44.3 &                    86.3 &           57.5 &           54.7 &     81.4 &               30.0 &        63.1 \\
Mis.-v0.1-7B & LLaMA-3.1-8B &                77.6 &            47.3 &                    87.3 &           58.2 &           56.5 &     83.1 &               30.8 &        65.0 \\
Mis.-v0.1-7B & LLaMA-3.1-70B &                77.9 &            49.1 &                    87.4 &           57.6 &           57.1 &     82.9 &               30.1 &        65.5 \\
\midrule
Mis.-v0.2-7B & Public data only &                76.7 &            46.8 &                    87.6 &           59.0 &           58.7 &     83.7 &               30.1 &        65.4 \\
Mis.-v0.2-7B & LLaMA-3.1-8B &                78.7 &            49.8 &                    87.9 &           58.5 &           59.1 &     83.7 &               30.7 &        66.6 \\ 
Mis.-v0.2-7B & LLaMA-3.1-70B & 78.3 & 50.5 & 88.2 & 60.0 & 58.2 & 85.7 & 31.3 & 66.9 \\
\bottomrule
\end{tabular}
\caption{Full MTEB evaluations for each model.}
\label{app_tab:full_mteb}
\end{table*}

\subsection{Omitted influence plots}

\begin{figure*}[ht!]
    \centering
    
    \begin{subfigure}[b]{0.48\textwidth}
        \centering
        \includegraphics[width=\textwidth]{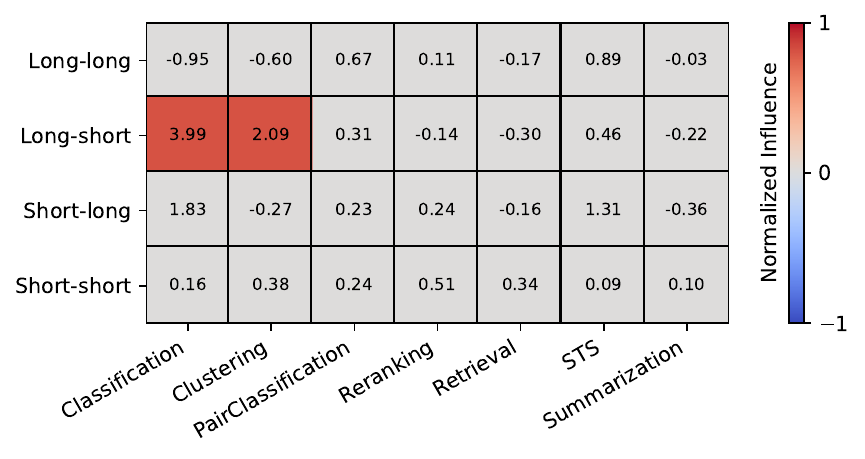}
        \caption{Mistral-v0.2-7B cross influence with LLaMA-3.1-8B.}
        \label{app_fig:cross-influence-mistral-7b-appendix}
    \end{subfigure}
    \hfill
    \begin{subfigure}[b]{0.48\textwidth}
        \centering
        \includegraphics[width=\textwidth]{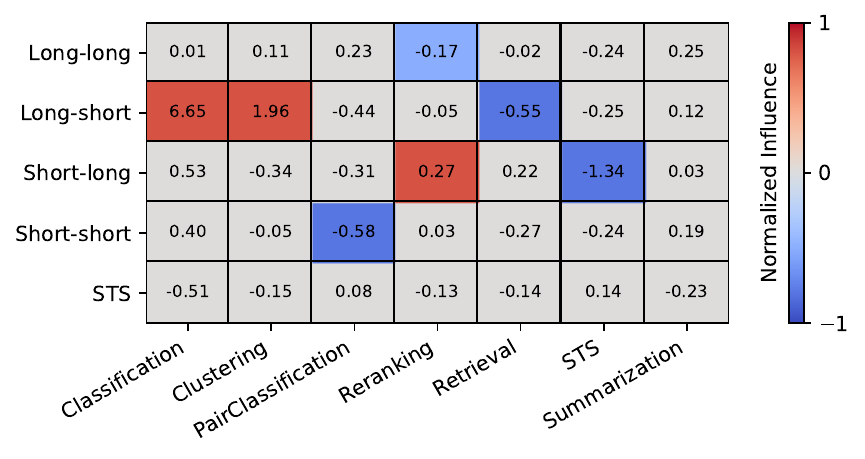}
        \caption{Mistral-v0.2-7B cross influence with LLaMA-3.1-70B.}
        \label{app_fig:cross-influence-mistral-70b-appendix}
    \end{subfigure}
    
    \vspace{0.5cm} %

    \begin{subfigure}[b]{0.48\textwidth}
        \centering
        \includegraphics[width=\textwidth]{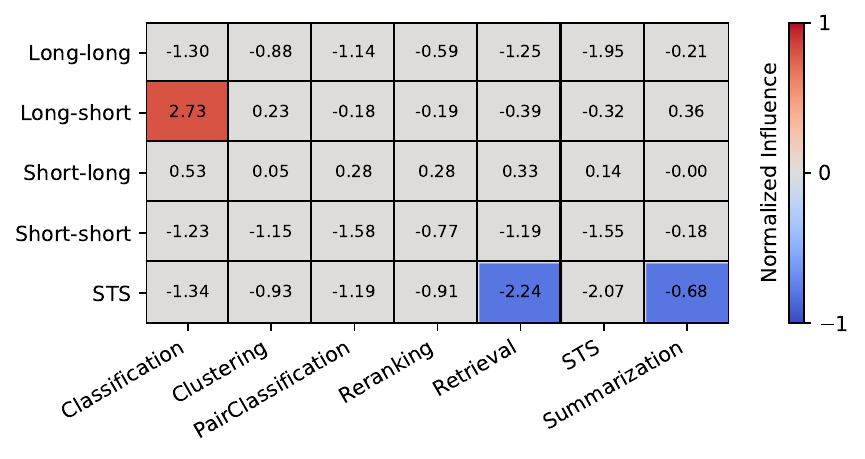}
        \caption{Qwen2-1.5B cross influence with LLaMA-3.1-8B.}
        \label{app_fig:cross-influence-qwen-7b-appendix}
    \end{subfigure}
    \hfill
    \begin{subfigure}[b]{0.48\textwidth}
        \centering
        \includegraphics[width=\textwidth]{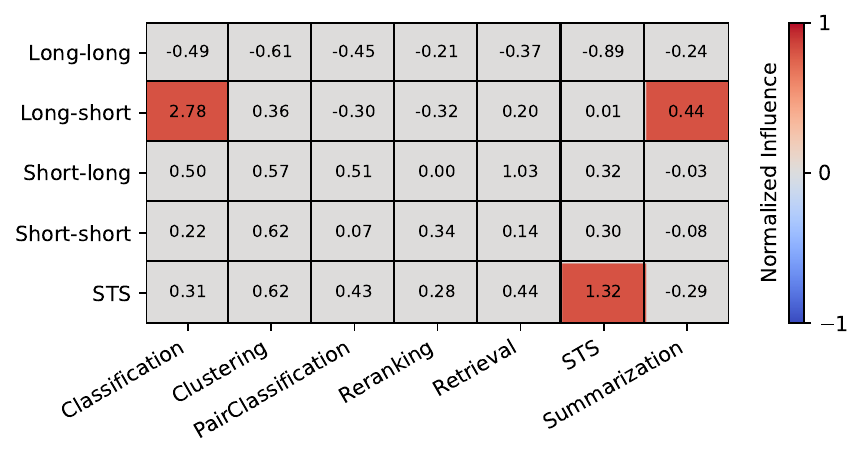}
        \caption{Qwen2-1.5B cross influence with LLaMA-3.1-70B.}
        \label{app_fig:cross-influence-qwen-70b-appendix}
    \end{subfigure}

    \caption{Comparison of cross-influence results for different models and synthetic datasets.}
    \label{fig:cross-influence-comparison}
\end{figure*}

\begin{figure*}[ht!]
    \centering
    
    \begin{subfigure}[b]{0.48\textwidth}
        \centering
        \includegraphics[width=\textwidth]{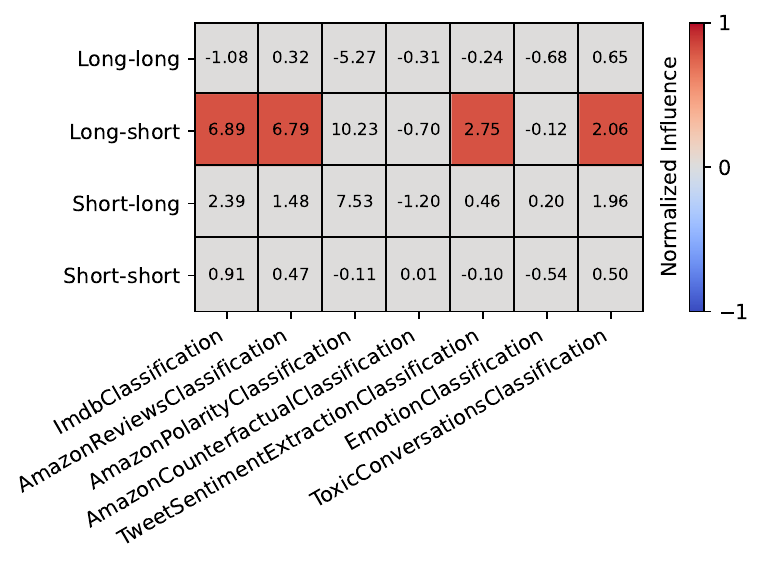}
        \caption{Classification}
        \label{app_fig:cross-influence-mistral-7b-appendix-Classification}
    \end{subfigure}
    \hfill
    \begin{subfigure}[b]{0.48\textwidth}
        \centering
        \includegraphics[width=\textwidth]{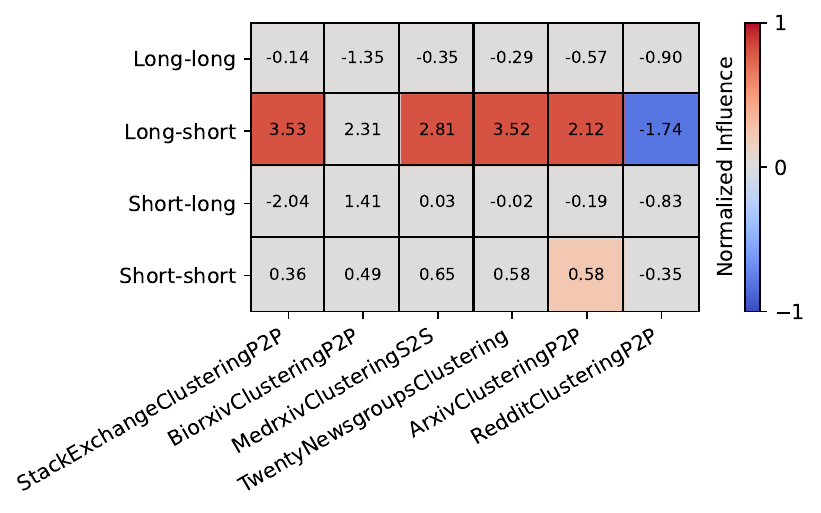}
        \caption{Clustering}
        \label{app_fig:cross-influence-mistral-7b-appendix-Clustering}
    \end{subfigure}
    
    \vspace{0.5cm}
    
    \begin{subfigure}[b]{0.48\textwidth}
        \centering
        \includegraphics[width=\textwidth]{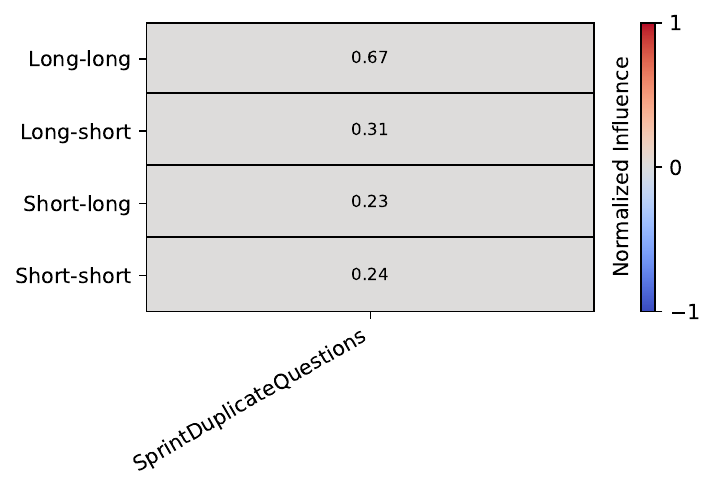}
        \caption{Pair Classification}
        \label{app_fig:cross-influence-mistral-7b-appendix-PairClassification}
    \end{subfigure}
    \hfill
    \begin{subfigure}[b]{0.48\textwidth}
        \centering
        \includegraphics[width=\textwidth]{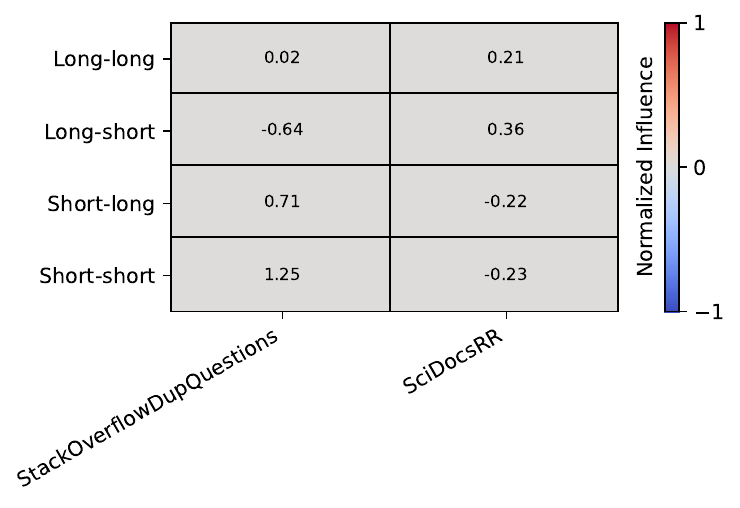}
        \caption{Reranking}
        \label{app_fig:cross-influence-mistral-7b-appendix-Reranking}
    \end{subfigure}

    \vspace{0.5cm} %

    \begin{subfigure}[b]{0.48\textwidth}
        \centering
        \includegraphics[width=\textwidth]{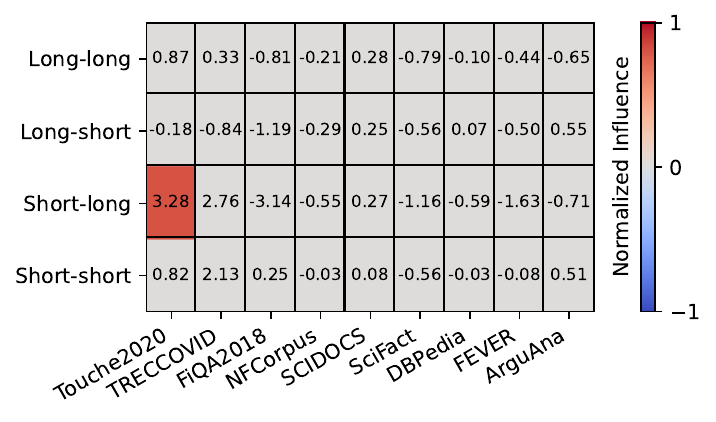}
        \caption{Retrieval}
        \label{app_fig:cross-influence-mistral-7b-appendix-Retrieval}
    \end{subfigure}
    \hfill
    \begin{subfigure}[b]{0.48\textwidth}
        \centering
        \includegraphics[width=\textwidth]{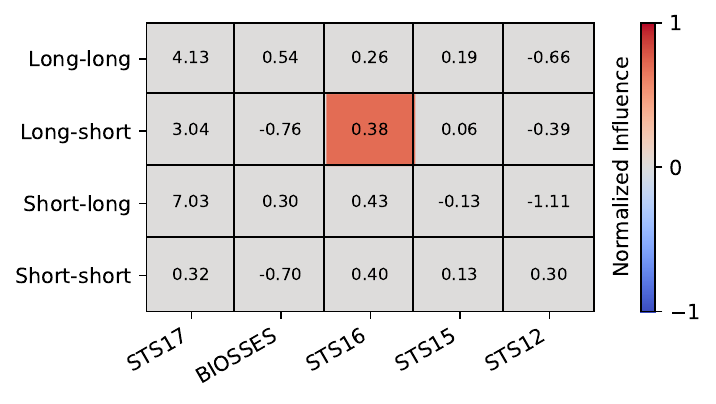}
        \caption{STS}
        \label{app_fig:cross-influence-mistral-7b-appendix-STS}
    \end{subfigure}
    
    \vspace{0.5cm}
    
    \begin{subfigure}[b]{0.48\textwidth} %
        \centering
        \includegraphics[width=\textwidth]{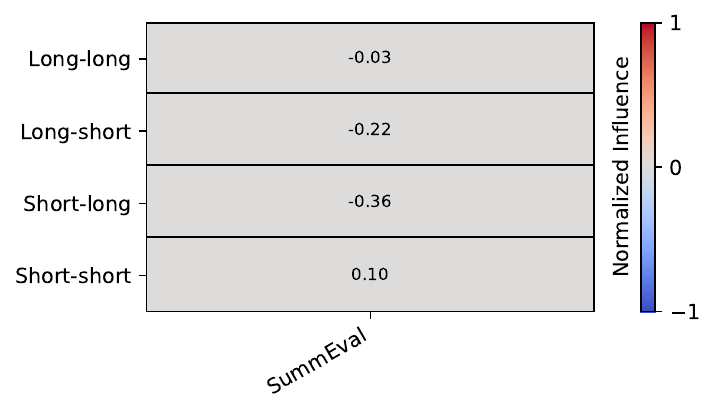}
        \caption{Summarization}
        \label{app_fig:cross-influence-mistral-7b-appendix-Summarization}
    \end{subfigure}

    \caption{Detailed breakdown of Mistral-v0.2-7B influence on various tasks with LLaMA-3.1-8B synthetic data.}
    \label{fig:cross-influence-mistral-summary}
\end{figure*}

\begin{figure*}[ht!]
    \centering
    
    \begin{subfigure}[b]{0.48\textwidth}
        \centering
        \includegraphics[width=\textwidth]{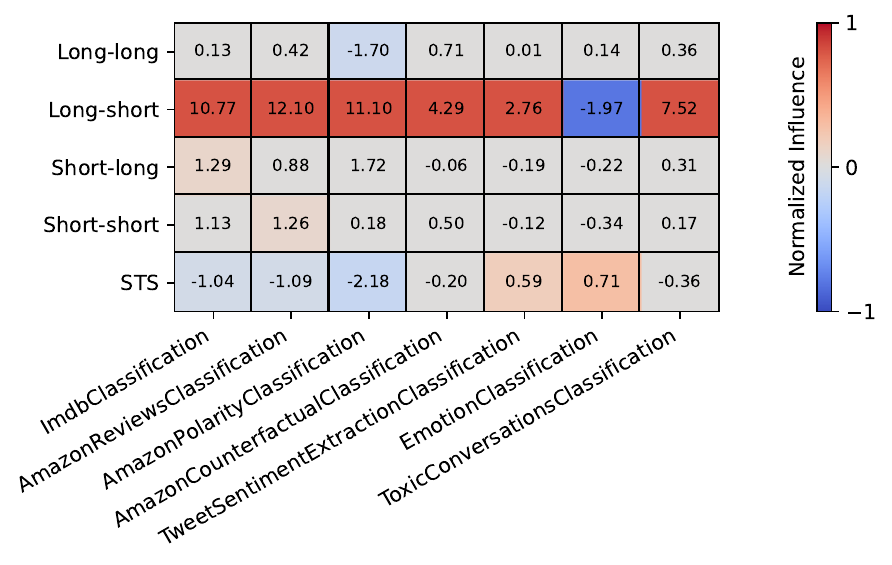}
        \caption{Classification}
        \label{app_fig:cross-influence-mistral-70b-appendix-Classification}
    \end{subfigure}
    \hfill
    \begin{subfigure}[b]{0.48\textwidth}
        \centering
        \includegraphics[width=\textwidth]{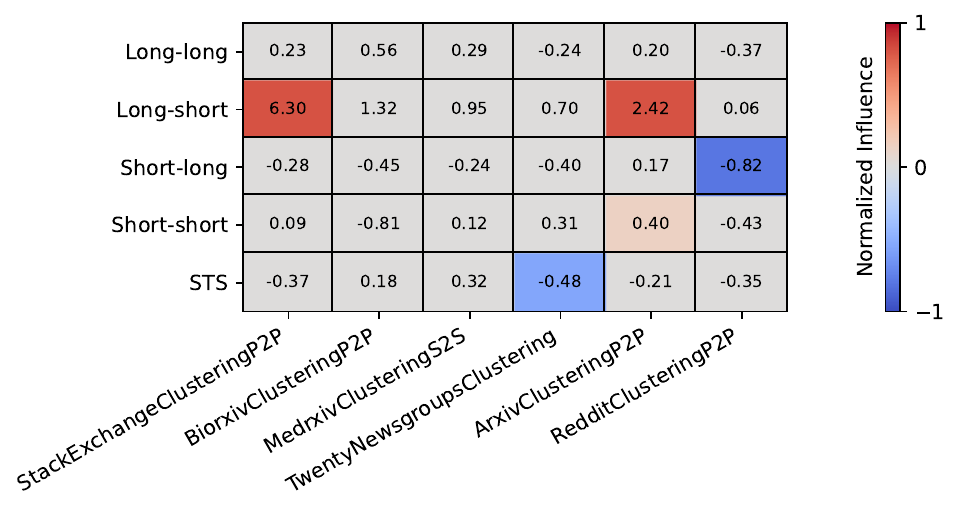}
        \caption{Clustering}
        \label{app_fig:cross-influence-mistral-70b-appendix-Clustering}
    \end{subfigure}

    \vspace{0.5cm} %

    \begin{subfigure}[b]{0.48\textwidth}
        \centering
        \includegraphics[width=\textwidth]{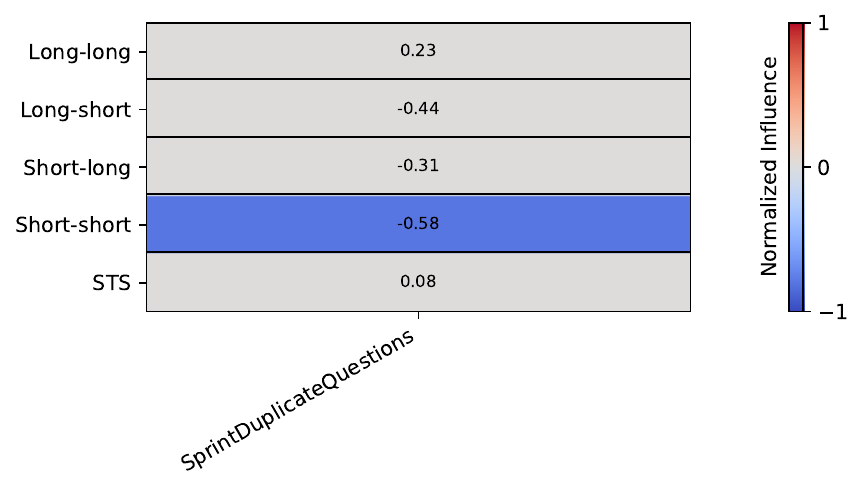}
        \caption{Pair Classification}
        \label{app_fig:cross-influence-mistral-70b-appendix-PairClassification}
    \end{subfigure}
    \hfill
    \begin{subfigure}[b]{0.48\textwidth}
        \centering
        \includegraphics[width=\textwidth]{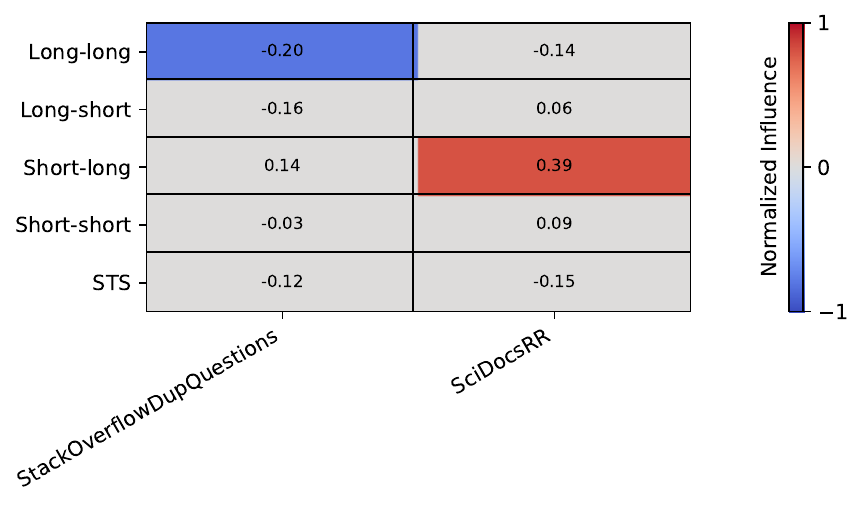}
        \caption{Reranking}
        \label{app_fig:cross-influence-mistral-70b-appendix-Reranking}
    \end{subfigure}

    \vspace{0.5cm}

    \begin{subfigure}[b]{0.48\textwidth}
        \centering
        \includegraphics[width=\textwidth]{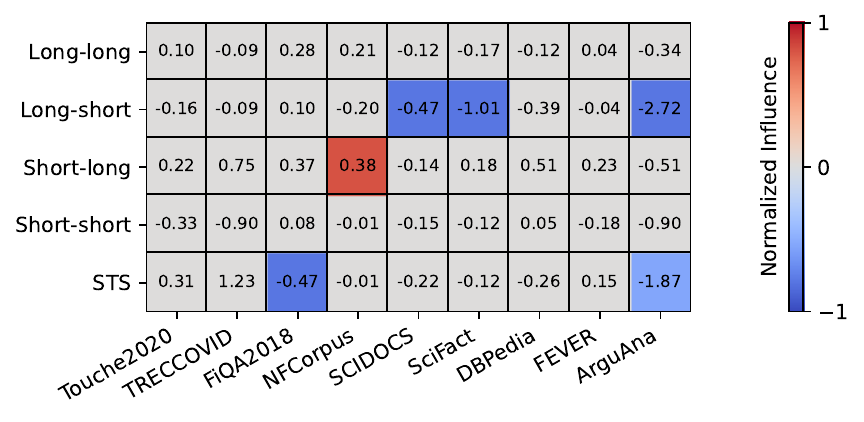}
        \caption{Retrieval}
        \label{app_fig:cross-influence-mistral-70b-appendix-Retrieval}
    \end{subfigure}
    \hfill
    \begin{subfigure}[b]{0.48\textwidth}
        \centering
        \includegraphics[width=\textwidth]{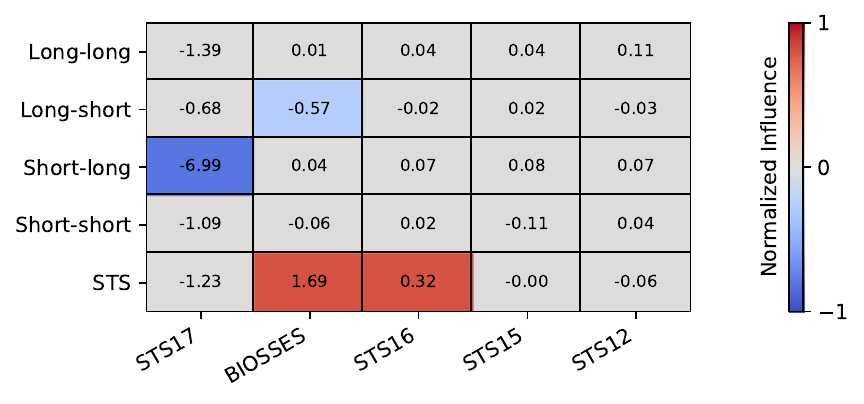}
        \caption{STS}
        \label{app_fig:cross-influence-mistral-70b-appendix-STS}
    \end{subfigure}

    \vspace{0.5cm}

    \begin{subfigure}[b]{0.48\textwidth}
        \centering
        \includegraphics[width=\textwidth]{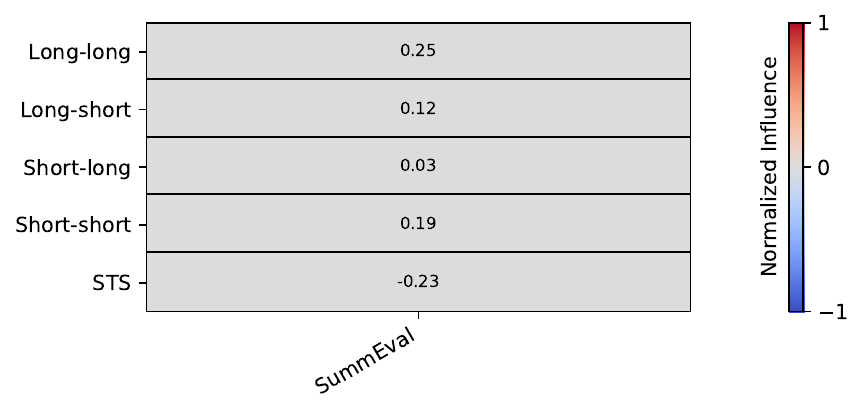}
        \caption{Summarization}
        \label{app_fig:cross-influence-mistral-70b-appendix-Summarization}
    \end{subfigure}

    \caption{Detailed breakdown of Mistral-v0.2-7B influence on various tasks with LLaMA-3.1-70B synthetic data.}
    \label{fig:cross-influence-mistral-70b-summary}
\end{figure*}

\begin{figure*}[ht!]
    \centering
    
    \begin{subfigure}[b]{0.48\textwidth}
        \centering
        \includegraphics[width=\textwidth]{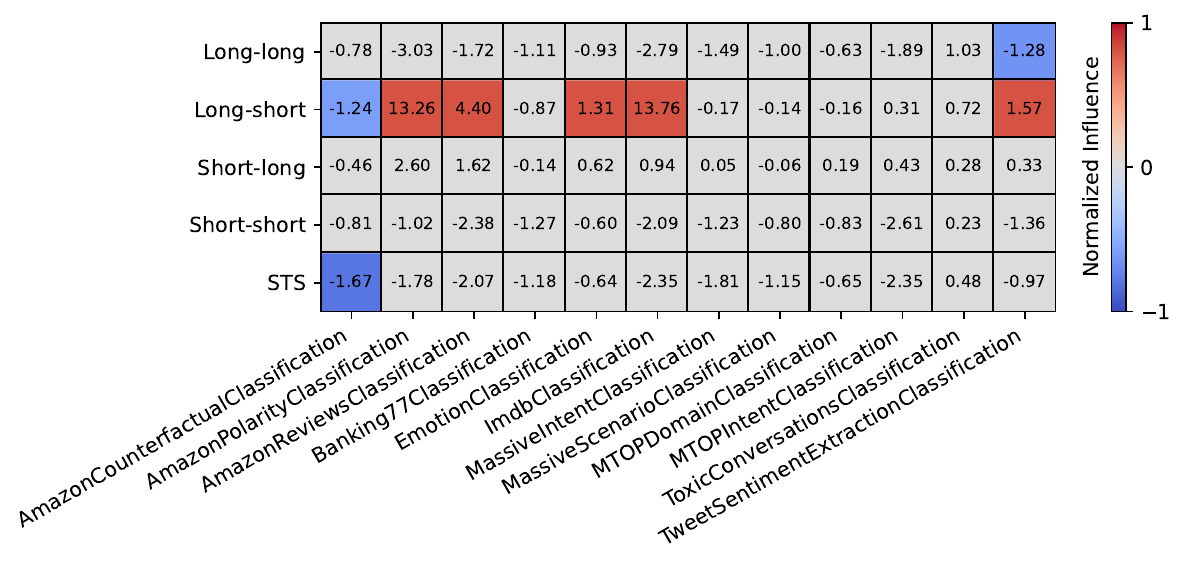}
        \caption{Classification}
        \label{app_fig:cross-influence-qwen-7b-appendix-Classification}
    \end{subfigure}
    \hfill
    \begin{subfigure}[b]{0.48\textwidth}
        \centering
        \includegraphics[width=\textwidth]{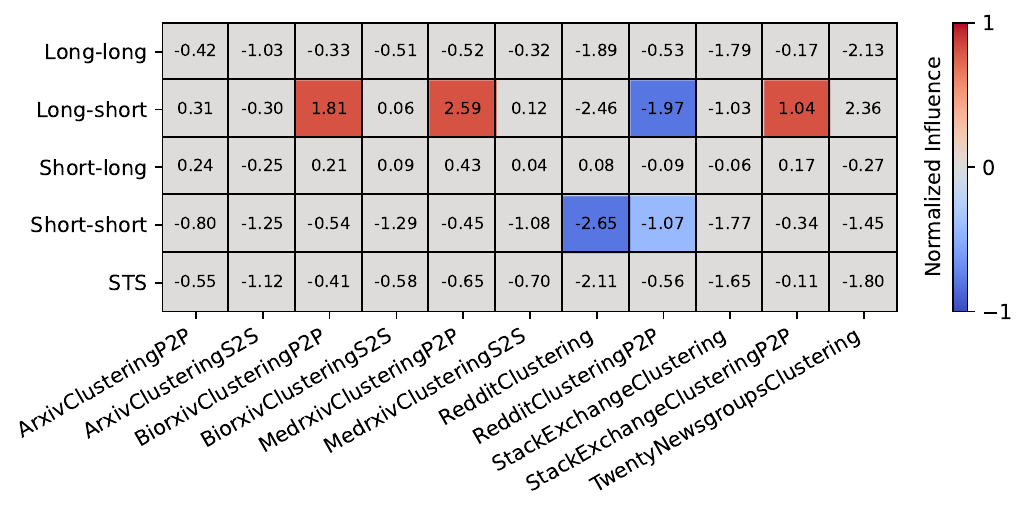}
        \caption{Clustering}
        \label{app_fig:cross-influence-qwen-7b-appendix-Clustering}
    \end{subfigure}

    \vspace{0.5cm} %

    \begin{subfigure}[b]{0.48\textwidth}
        \centering
        \includegraphics[width=\textwidth]{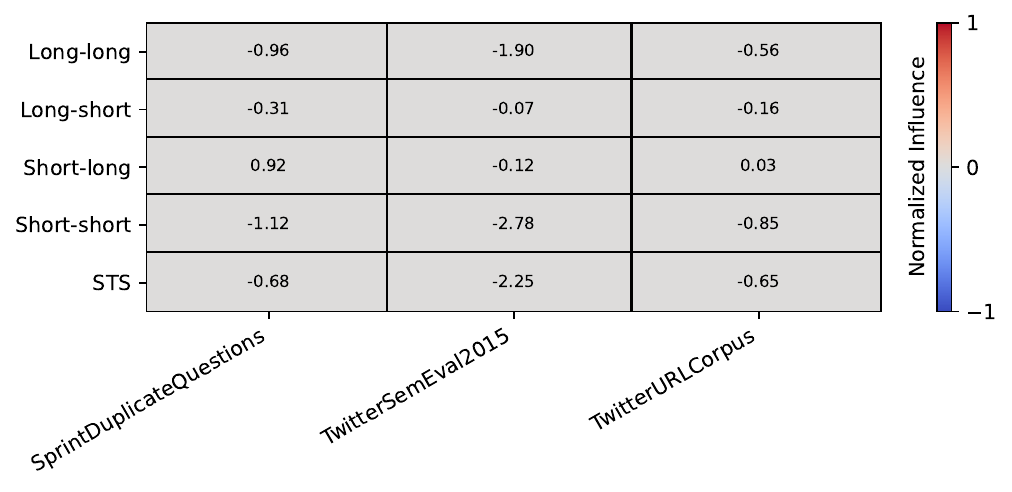}
        \caption{Pair Classification}
        \label{app_fig:cross-influence-qwen-7b-appendix-PairClassification}
    \end{subfigure}
    \hfill
    \begin{subfigure}[b]{0.48\textwidth}
        \centering
        \includegraphics[width=\textwidth]{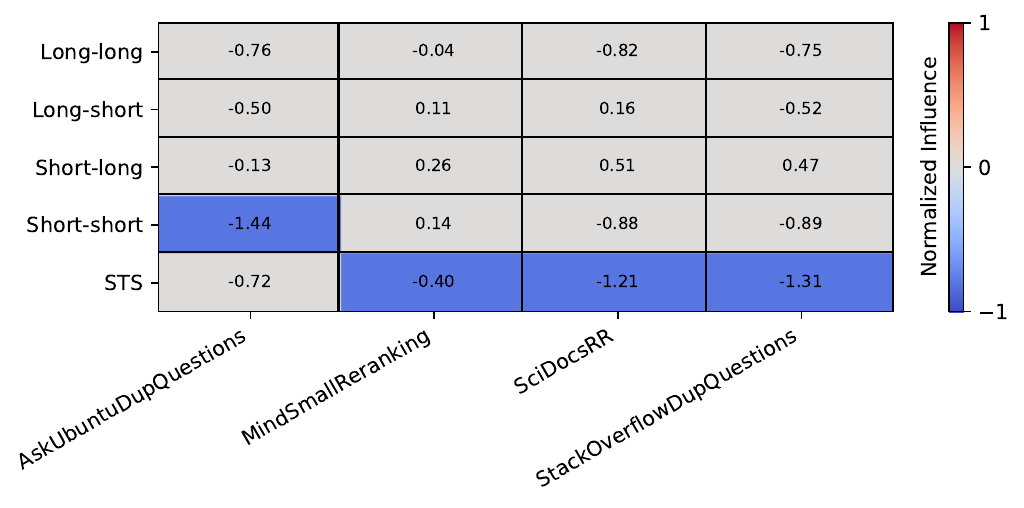}
        \caption{Reranking}
        \label{app_fig:cross-influence-qwen-7b-appendix-Reranking}
    \end{subfigure}

    \vspace{0.5cm}

    \begin{subfigure}[b]{0.48\textwidth}
        \centering
        \includegraphics[width=\textwidth]{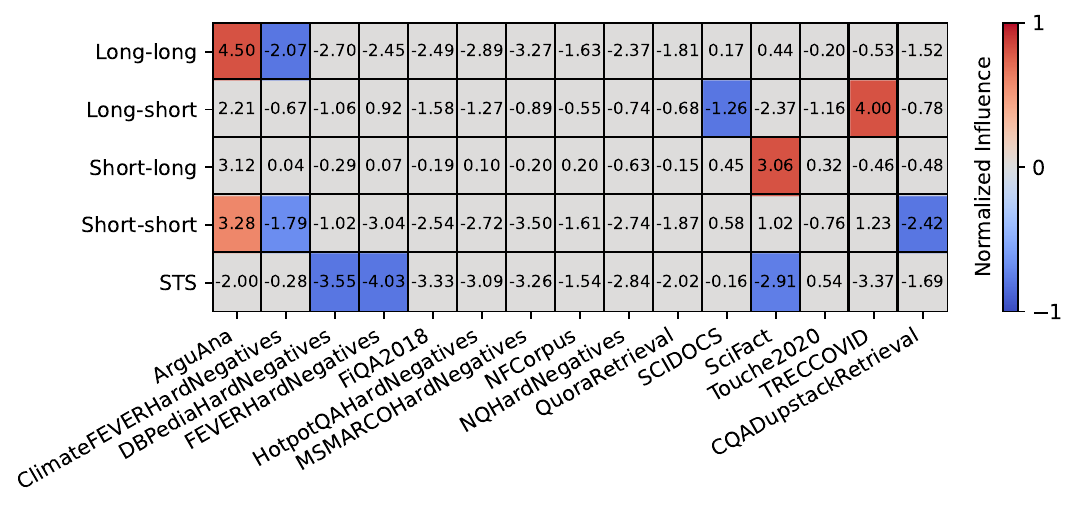}
        \caption{Retrieval}
        \label{app_fig:cross-influence-qwen-7b-appendix-Retrieval}
    \end{subfigure}
    \hfill
    \begin{subfigure}[b]{0.48\textwidth}
        \centering
        \includegraphics[width=\textwidth]{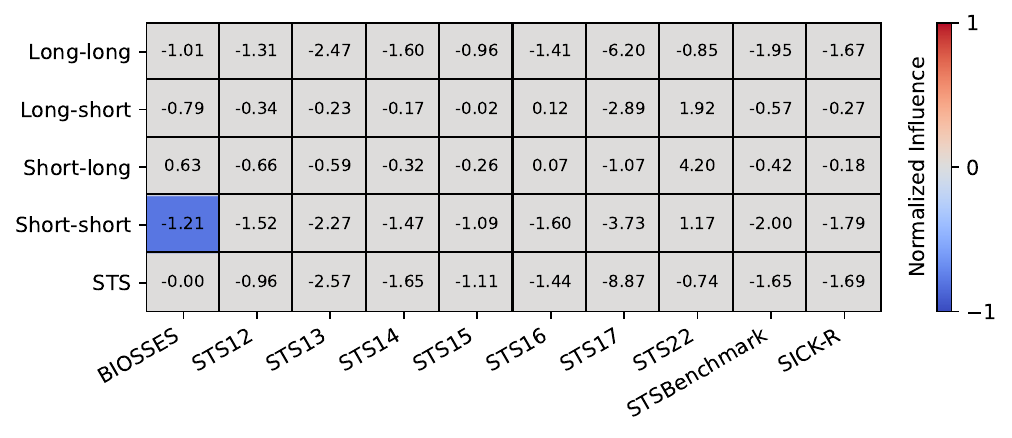}
        \caption{STS}
        \label{app_fig:cross-influence-qwen-7b-appendix-STS}
    \end{subfigure}

    \vspace{0.5cm}

    \begin{subfigure}[b]{0.48\textwidth}
        \centering
        \includegraphics[width=\textwidth]{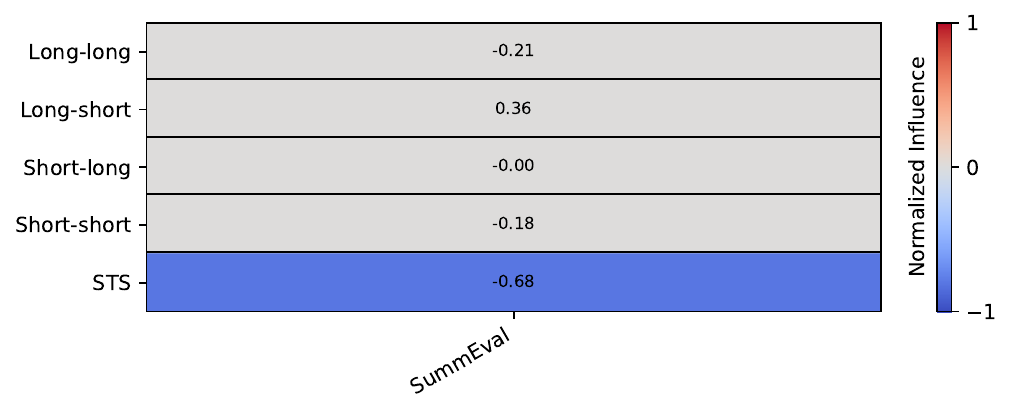}
        \caption{Summarization}
        \label{app_fig:cross-influence-qwen-7b-appendix-Summarization}
    \end{subfigure}

    \caption{Detailed breakdown of Qwen2-1.5B influence on various tasks with LLaMA-3.1-8B synthetic data.}
    \label{fig:cross-influence-qwen-7b-summary}
\end{figure*}

\begin{figure*}[ht!]
    \centering
    
    \begin{subfigure}[b]{0.48\textwidth}
        \centering
        \includegraphics[width=\textwidth]{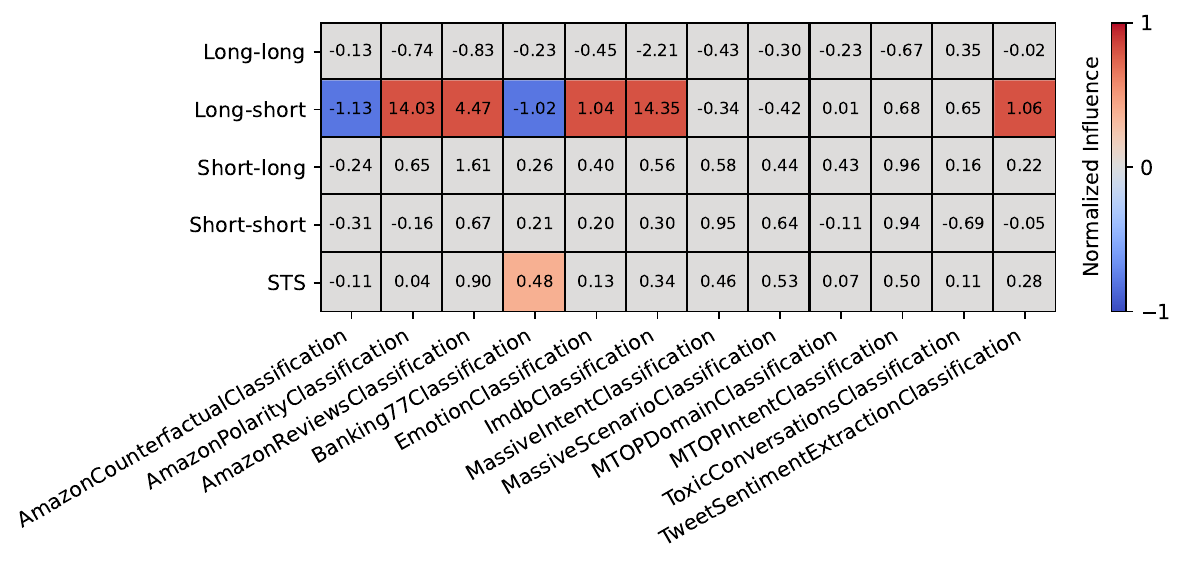}
        \caption{Classification}
        \label{app_fig:cross-influence-qwen-70b-appendix-Classification}
    \end{subfigure}
    \hfill
    \begin{subfigure}[b]{0.48\textwidth}
        \centering
        \includegraphics[width=\textwidth]{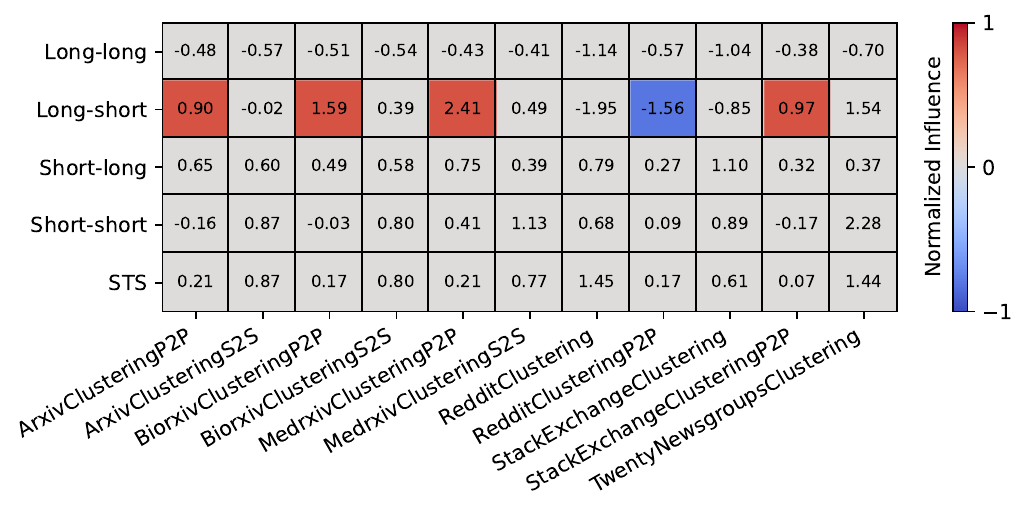}
        \caption{Clustering}
        \label{app_fig:cross-influence-qwen-70b-appendix-Clustering}
    \end{subfigure}

    \vspace{0.5cm} %

    \begin{subfigure}[b]{0.48\textwidth}
        \centering
        \includegraphics[width=\textwidth]{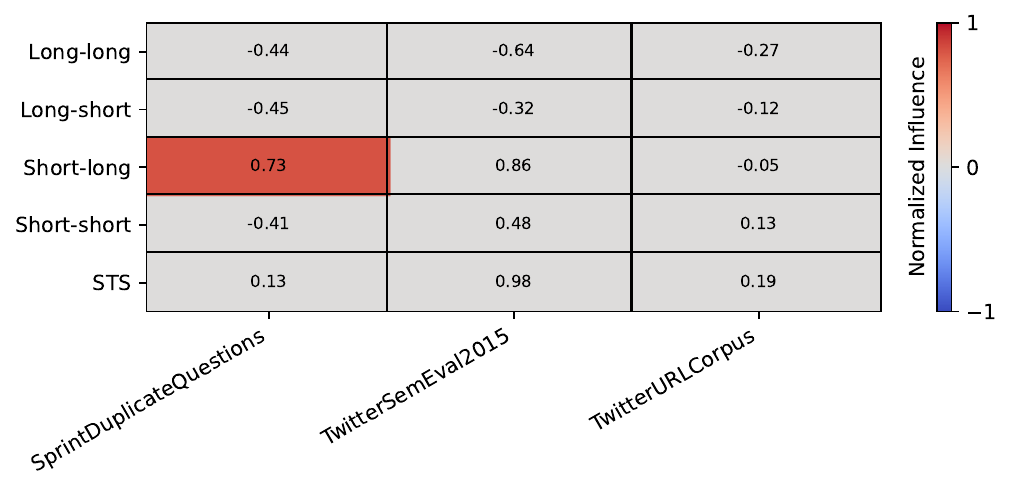}
        \caption{Pair Classification}
        \label{app_fig:cross-influence-qwen-70b-appendix-PairClassification}
    \end{subfigure}
    \hfill
    \begin{subfigure}[b]{0.48\textwidth}
        \centering
        \includegraphics[width=\textwidth]{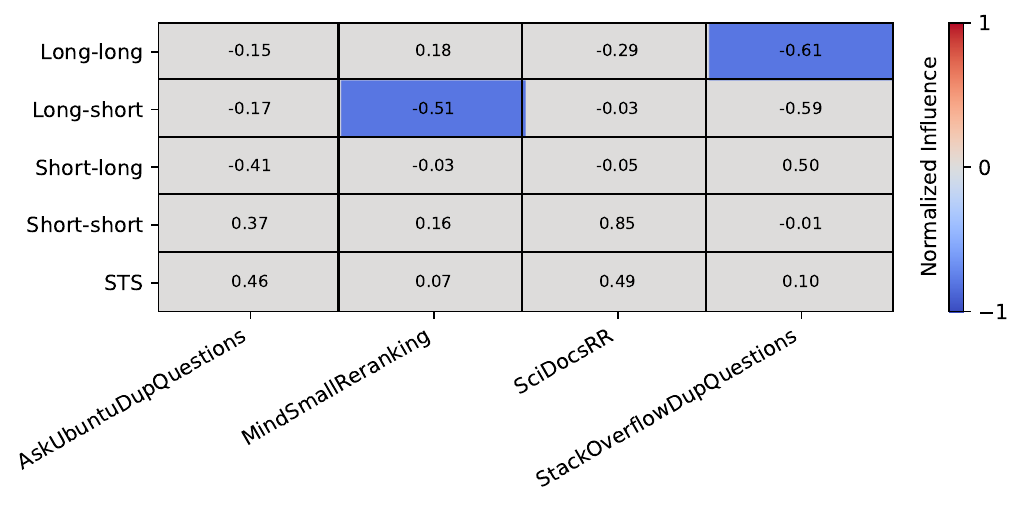}
        \caption{Reranking}
        \label{app_fig:cross-influence-qwen-70b-appendix-Reranking}
    \end{subfigure}

    \vspace{0.5cm}

    \begin{subfigure}[b]{0.48\textwidth}
        \centering
        \includegraphics[width=\textwidth]{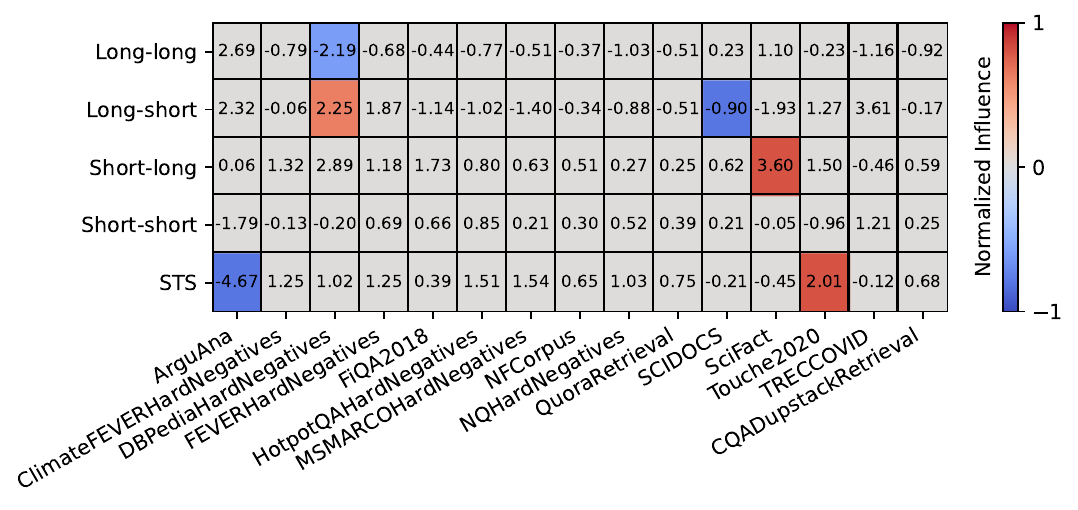}
        \caption{Retrieval}
        \label{app_fig:cross-influence-qwen-70b-appendix-Retrieval}
    \end{subfigure}
    \hfill
    \begin{subfigure}[b]{0.48\textwidth}
        \centering
        \includegraphics[width=\textwidth]{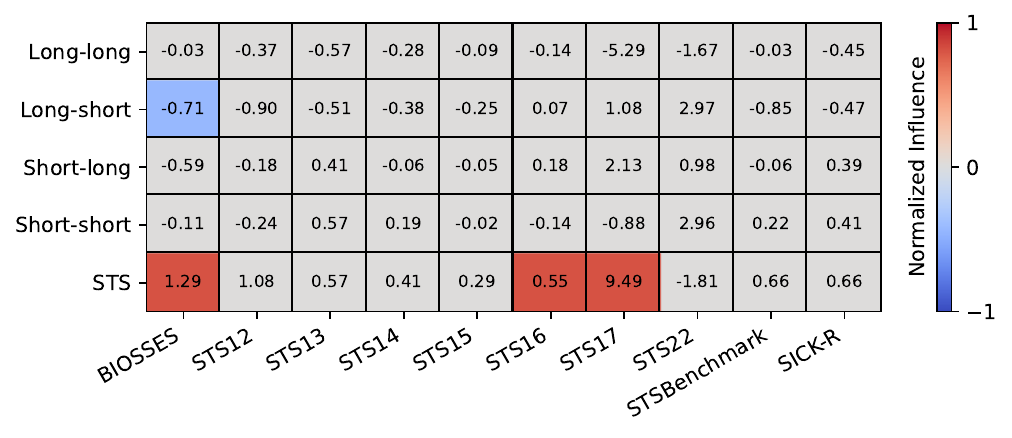}
        \caption{STS}
        \label{app_fig:cross-influence-qwen-70b-appendix-STS}
    \end{subfigure}

    \vspace{0.5cm}

    \begin{subfigure}[b]{0.48\textwidth}
        \centering
        \includegraphics[width=\textwidth]{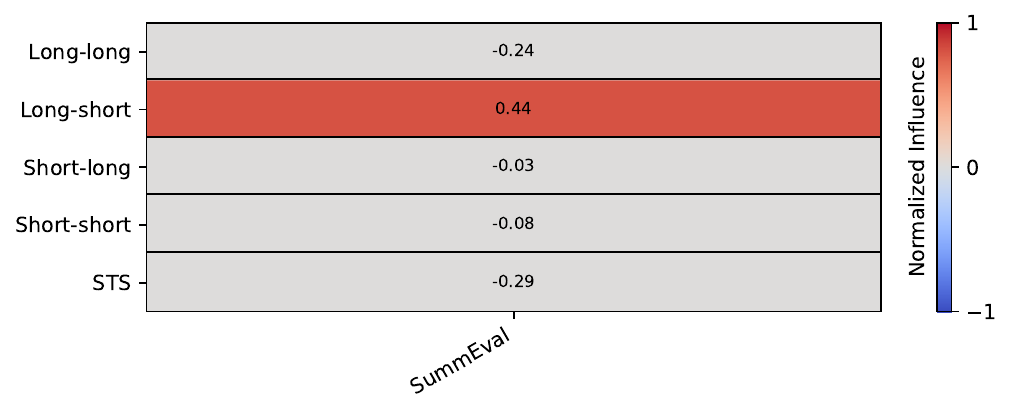}
        \caption{Summarization}
        \label{app_fig:cross-influence-qwen-70b-appendix-Summarization}
    \end{subfigure}

    \caption{Detailed breakdown of Qwen2-1.5B influence on various tasks with LLaMA-3.1-70B synthetic data.}
    \label{fig:cross-influence-qwen-70b-summary}
\end{figure*}

We extend Figures~\ref{fig:category-ablations} and~\ref{fig:dataset-ablations} with the results from all four settings: Mistral-v0.2-7B and Qwen2-1.5B base models, trained on LLaMA-3.1-8B or LLaMA-3.1-70B synthetic data. We plot the results in Figures~\ref{fig:cross-influence-comparison},~\ref{fig:cross-influence-mistral-summary},~\ref{fig:cross-influence-mistral-70b-summary},~\ref{fig:cross-influence-qwen-7b-summary},~ and~\ref{fig:cross-influence-qwen-70b-summary}. In general, we observe similar trends from the main paper.

\newpage
\begin{table*}[t]
    \centering
    \small
    \begin{tabular}{@{}p{\linewidth}@{}}
    \toprule
    \textbf{Synthetic data category:} short-short \\ \midrule
    \textbf{Instruction:} Identify a famous painting from a brief description. \\ \midrule
    \multicolumn{1}{c}{\textit{Llama-3.1-70B}} \\ \midrule
    \textbf{Generated data:} \{\\
    "query": "Painting of a woman with a mysterious smile", \\
    "positive": "The Mona Lisa by Leonardo da Vinci", \\
    "negative": "Claude Monet" \\
    \} \\ \midrule
    \multicolumn{1}{c}{\textit{Llama-3.1-8B}} \\ \midrule
    \textbf{Generated data:} \{\\
    "query": "a woman with a clock", \\
    "positive": "Girl with a Pearl Earring by Johannes Vermeer", \\
    "negative": "a toy that runs using a spring" \\
    \} \\
    \bottomrule
    \end{tabular}
    \caption{Sample from \textit{short-short} subgroup of the synthetic data.}
    \label{tab:app_synthetic_data_example_short_short}
    \end{table*}
    
    \begin{table*}[t]
    \centering
    \small
    \begin{tabular}{@{}p{\linewidth}@{}}
    \toprule
    \textbf{Synthetic data category:} short-long \\ \midrule
    \textbf{Instruction:} Search for documentaries about the effects of pollution on human health. \\ \midrule
    \multicolumn{1}{c}{\textit{Llama-3.1-70B}} \\ \midrule
    \textbf{Generated data:} \{\\
    "query": "Documentaries exploring the long-term health effects of microplastic consumption in human populations", \\
    "positive": "The Devil We Know is a 2018 documentary that examines the effects of plastics on human health and the environment. The film focuses on the history of plastics and their impact on human health, including the leaching of chemicals into food and water. ...\textit{\textbf{truncated for brevity}} .... While the film does not focus exclusively on microplastic consumption, it provides a comprehensive overview of the issue and its effects on human health. The documentary has been praised for its thorough research and engaging storytelling, and has won several awards for its contribution to the public understanding of environmental issues. The film is a valuable resource for anyone looking to understand the impact of plastics on human health, and the ways in which individuals and communities can work to reduce their exposure to these pollutants.", \\
    "negative": "Plastic Ocean is a 2016 documentary that explores the impact of plastic pollution on marine ecosystems. The film follows a team of researchers as they travel the world, documenting the effects of plastic waste on ocean life. ...\textit{\textbf{truncated for brevity}} .... The film has been praised for its visually stunning footage and its ability to raise awareness about the issue of plastic pollution." \\
    \} \\ \midrule
    \multicolumn{1}{c}{\textit{Llama-3.1-8B}} \\ \midrule
    \textbf{Generated data:} \{\\
    "query": "The effects of particulate matter on cardiovascular disease in low-income communities in urban areas with high levels of air pollution", \\
    "positive": "Air pollution is a significant public health concern worldwide, particularly in low-income communities. The World Health Organization (WHO) estimates that 9 out of 10 people worldwide breathe polluted air, resulting in over 7 million premature deaths annually. One of the primary pollutants responsible for this alarming rate is particulate matter (PM). ...\textit{\textbf{truncated for brevity}} ....To mitigate the effects of PM on cardiovascular disease, several strategies can be implemented. These include reducing emissions from vehicles, industrial sites, and power plants, increasing the use of renewable energy sources, and promoting green infrastructure such as green roofs and urban parks. Additionally, improving air quality monitoring and public education can help raise awareness about the risks associated with PM exposure. In conclusion, the effects of particulate matter on cardiovascular disease in low-income communities in urban areas with high levels of air pollution are a pressing public health concern. It is essential to address this issue through a combination of policy changes, technological innovations, and community-based interventions to reduce PM exposure and protect the health and well-being of vulnerable populations.", \\
    "negative": "The effects of air pollution on human health have been a topic of interest for many years. However, the relationship between particulate matter and cardiovascular disease is not as clear-cut as previously thought. While some studies have found a link between PM exposure and cardiovascular disease, others have found no significant association.  A 2019 study published in the Journal of Exposure Science \& Environmental Epidemiology found that exposure to PM2.5 was associated with an increased risk of cardiovascular disease in a cohort of older adults. However, this study had several limitations, including a small sample size ...\textit{\textbf{truncated for brevity}} .... In conclusion, while some studies have found a link between PM exposure and cardiovascular disease, the evidence is not conclusive, and more research is needed to fully understand the relationship between these two factors. It is essential to consider the complexity of the issue and the potential confounding variables when interpreting the results of studies on this topic." \\
    \} \\
    \bottomrule
    \end{tabular}
    \caption{Sample from \textit{short-long} subgroup of the synthetic data.}
    \label{tab:app_synthetic_data_example_short_long}
    \end{table*}
    
    \begin{table*}[t]
    \centering
    \small
    \begin{tabular}{@{}p{\linewidth}@{}}
    \toprule
    \textbf{Synthetic data category:} long-short \\ \midrule
    \textbf{Instruction:} Classify data breach notifications into high, medium, and low risk categories. \\ \midrule
    \multicolumn{1}{c}{\textit{Llama-3.1-70B}} \\ \midrule
    \textbf{Generated data:} \{\\
    "query": "A security incident has been reported at a prominent e-commerce platform, affecting the sensitive information of over 500,000 customers. According to the company's investigation, a vulnerability in their login system allowed unauthorized access to their database. While the company assures that no financial data was compromised, email addresses, phone numbers, and physical addresses were exposed. The incident occurred due to a misconfigured security update, which left the system open to attacks for a period of three weeks. The company is working closely with law enforcement agencies to identify the perpetrators and has implemented additional security measures to prevent similar incidents in the future. As a precautionary measure, all affected customers have been notified and advised to change their passwords immediately.", \\
    "positive": "high risk", \\
    "negative": "medium risk" \\
    \} \\ \midrule
    \multicolumn{1}{c}{\textit{Llama-3.1-8B}} \\ \midrule
    \textbf{Generated data:} \{\\
    "query": "A sophisticated phishing attack targeted a large corporation's IT department, resulting in the unauthorized access to sensitive customer data. The attackers used a combination of social engineering tactics and zero-day exploits to breach the company's network, ultimately gaining access to over 1 million customer records. The breach was discovered after a routine security audit revealed suspicious network activity. The affected data includes names, addresses, and payment card information. The company is taking immediate action to notify affected customers and is cooperating with law enforcement to investigate the incident.", \\
    "positive": "high", \\
    "negative": "medium" \\
    \} \\
    \bottomrule
    \end{tabular}
    \caption{Sample from \textit{long-short} subgroup of the synthetic data.}
    \label{tab:app_synthetic_data_example_long_short}
    \end{table*}
    
    \begin{table*}[t]
    \centering
    \small
    \begin{tabular}{@{}p{\linewidth}@{}}
    \toprule
    \textbf{Synthetic data category:} long-long \\ \midrule
    \textbf{Instruction:} Given comprehensive insights into the utility of meditation, locate opposing views arguing against its psychological benefits. \\ \midrule
    \multicolumn{1}{c}{\textit{Llama-3.1-70B}} \\ \midrule
    \textbf{Generated data:} \{\\
    "query": "While numerous studies have demonstrated the benefits of meditation for both physical and mental well-being, a growing body of research has begun to challenge the notion that meditation is a universally beneficial practice. Critics argue that meditation can have negative effects on certain individuals, particularly those with a history of trauma or mental health issues. For example, some studies have found that mindfulness meditation can exacerbate symptoms of anxiety and depression in individuals who are prone to rumination, or the tendency to repetitively think negative thoughts. Additionally, meditation can be used as a form of avoidance, allowing individuals to disengage from their problems rather than confronting and resolving them. Furthermore, the emphasis on individualism and self-improvement in Westernized forms of meditation can be alienating for individuals from collectivist cultures, who may view meditation as a selfish pursuit. Moreover, the commercialization of meditation has led to the proliferation of inexperienced and unqualified teachers, who may not be equipped to provide adequate guidance and support to students. In light of these findings, it is essential to approach the practice of meditation with a critical and nuanced perspective, recognizing both its potential benefits and limitations.", \\
    "positive": "The benefits of meditation have been extensively researched and documented, with studies demonstrating its positive impact on both physical and mental health. ...\textit{\textbf{truncated for brevity}} .... As the scientific evidence supporting the benefits of meditation continues to grow, it is becoming increasingly clear that this ancient practice has the potential to improve overall well-being and quality of life.", \\
    "negative": "A recent study published in the Journal of Clinical Psychology found that participants in an MBSR program exhibited significant reductions in symptoms of anxiety and depression, as well as improved mood and quality of life. The study, which was conducted at a large urban medical center, included 100 participants who were randomly assigned to either an MBSR group or a wait-list control group. ...\textit{\textbf{truncated for brevity}} .... Participants in the wait-list control group did not receive any mindfulness training, but were given the option to participate in the MBSR program after the study was completed. The results of the study showed that participants in the MBSR group exhibited significant reductions in pain and improved mood, compared to participants in the wait-list control group. The study's authors note that further research is needed to fully understand the mechanisms by which MBSR programs achieve their benefits for individuals with chronic pain, but suggest that the program may be a useful tool for improving pain management outcomes." \\
    \} \\ \midrule
    \multicolumn{1}{c}{\textit{Llama-3.1-8B}} \\ \midrule
    \textbf{Generated data:} \{\\
    "query": "The effectiveness of meditation in reducing stress and anxiety has been extensively documented in various scientific studies. Meditation has been shown to decrease the production of stress hormones, such as cortisol, and increase the production of neurotransmitters that promote feelings of relaxation and well-being. Regular meditation practice has also been linked to improved sleep quality, reduced blood pressure, and a stronger immune system. Furthermore, meditation has been used as a tool for managing chronic pain, depression, and anxiety disorders. The benefits of meditation are not limited to physical health; it has also been shown to improve cognitive functioning, including attention, memory, and problem-solving skills. Meditation can be practiced in various forms, including mindfulness meditation, loving-kindness meditation, and transcendental meditation. The scientific community has extensively studied the effects of meditation, and the evidence suggests that it is a valuable tool for promoting overall well-being and reducing stress and anxiety. What are some opposing views that argue against the psychological benefits of meditation?", \\
    "positive": "Meditation has been widely accepted as a valuable tool for promoting mental and physical well-being. Research has shown that regular meditation practice can reduce stress and anxiety, improve sleep quality, and even alleviate symptoms of depression and anxiety disorders....\textit{\textbf{truncated for brevity}} .... The scientific community has extensively studied the effects of meditation, and the evidence suggests that it is a valuable tool for promoting overall well-being and reducing stress and anxiety. Regular meditation practice has been shown to have a positive impact on both physical and mental health, making it an important component of a healthy lifestyle.", \\
    "negative": "While the benefits of prioritizing mental health are well-documented, there are also potential drawbacks to consider. For example, a study published in the Journal of Behavioral Addictions found that excessive exercise and physical activity can actually exacerbate anxiety and depression in some individuals (Kvam et al., 2016). ...\textit{\textbf{truncated for brevity}} .... Moreover, some researchers have argued that the emphasis on individual responsibility for mental health can be problematic, as it can lead to stigma and shame for those who are struggling (Slaby, 2002). By taking a more nuanced approach to mental health, one that acknowledges the complexities and limitations of the current understanding, we can work towards creating a more comprehensive and effective approach to mental health support." \\
    \} \\
    \bottomrule
    \end{tabular}
    \caption{Sample from \textit{long-long} subgroup of the synthetic data.}
    \label{tab:app_synthetic_data_example_long_long}
    \end{table*}

\end{document}